\documentclass[lettersize,journal]{IEEEtran}
\pdfoutput=1
\usepackage{amsmath,amsfonts}
\usepackage{algorithmic}
\usepackage{algorithm}
\makeatletter
\newcommand{\removelatexerror}{\let\@latex@error\@gobble}
\makeatother
\usepackage{array}
\usepackage[caption=false,font=normalsize,labelfont=sf,textfont=sf]{subfig}
\usepackage{textcomp}
\usepackage{stfloats}
\usepackage{url}
\usepackage{verbatim}
\usepackage{graphicx}
\usepackage{cite}
\usepackage{color}
\usepackage{dsfont}
\usepackage{multirow}
\begin{document}

\title{Spatiotemporal Decouple-and-Squeeze Contrastive Learning for Semi-Supervised Skeleton-based Action Recognition}

\author{Binqian~Xu, Xiangbo~Shu, \textit{Senior Member, IEEE}
\thanks{\em B. Xu, and X. Shu are with the School of Computer Science and Engineering, Nanjing University of Science and Technology, Nanjing 210094, China. E-mail: xubinq11@gmail.com, shuxb@njust.edu.cn. (Corresponding author: Xiangbo Shu) (B.~Xu and X. Shu are co-first authors)}}

\markboth{Submission~of~IEEE~TRANSACTIONS~ON~NEURAL~NETWORKS~AND~LEARNING~SYSTEMS, 2022}%
{Submission~of~IEEE~TRANSACTIONS~ON~NEURAL~NETWORKS~AND~LEARNING~SYSTEMS, 2022}


\maketitle

\begin{abstract}
Contrastive learning has been successfully leveraged to learn action representations for addressing the problem of semi-supervised skeleton-based action recognition. However, most contrastive learning-based methods only contrast global features mixing spatiotemporal information, which confuses the spatial- and temporal-specific information reflecting different semantic at the frame level and joint level. Thus, we propose a novel Spatiotemporal Decouple-and-Squeeze Contrastive Learning (SDS-CL) framework to comprehensively learn more abundant representations of skeleton-based actions by jointly contrasting spatial-squeezing features, temporal-squeezing features, and global features. In SDS-CL, we design a new Spatiotemporal-decoupling Intra-Inter Attention (SIIA) mechanism to obtain the spatiotemporal-decoupling attentive features for capturing spatiotemporal specific information by calculating spatial- and temporal-decoupling intra-attention maps among joint/motion features, as well as spatial- and
temporal-decoupling inter-attention maps between joint and motion features. Moreover, we present a new Spatial-squeezing Temporal-contrasting Loss (STL), a new Temporal-squeezing Spatial-contrasting Loss (TSL), and the  Global-contrasting Loss (GL) to contrast the spatial-squeezing joint and motion features at the frame level, temporal-squeezing joint and motion features at the joint level, as well as global joint and motion features at the skeleton level. Extensive experimental results on four public datasets show that the proposed SDS-CL achieves performance gains compared with other competitive methods.
\end{abstract}


\section{Introduction}
\IEEEPARstart{H}{uman} action recognition is an attractive task in the computer vision area and has been studied extensively in the past decade due to its wide applications in video retrieval, video surveillance, virtual reality, and so on~\cite{weinland2011survey,poppe2010survey,shu2019hierarchical,shu2020host,shu2022expansion,tang2019coherence,liu2019hidden,yu2021searching}. Recently, skeletal data consisting of 3D coordinates of joints has become increasingly popular in the human action recognition task because it is a compact and high-level representation of action, which has strong adaptability to the human body scales, camera viewpoints, and dynamic circumstances~\cite{kim2017interpretable,li2018independently,shu2022spatiotemporal,du2015hierarchical,zhang2017view,li2018co,ke2017new, yan2018spatial,tang2018deep,liu20193d,liu20203d}. Moreover, skeletal data coming from depth sensors or pose estimation algorithms is more advantageous in computation and storage~\cite{li2018co,yan2017skeleton}.

\begin{figure}
    \centering
    \includegraphics[width=3.4in]{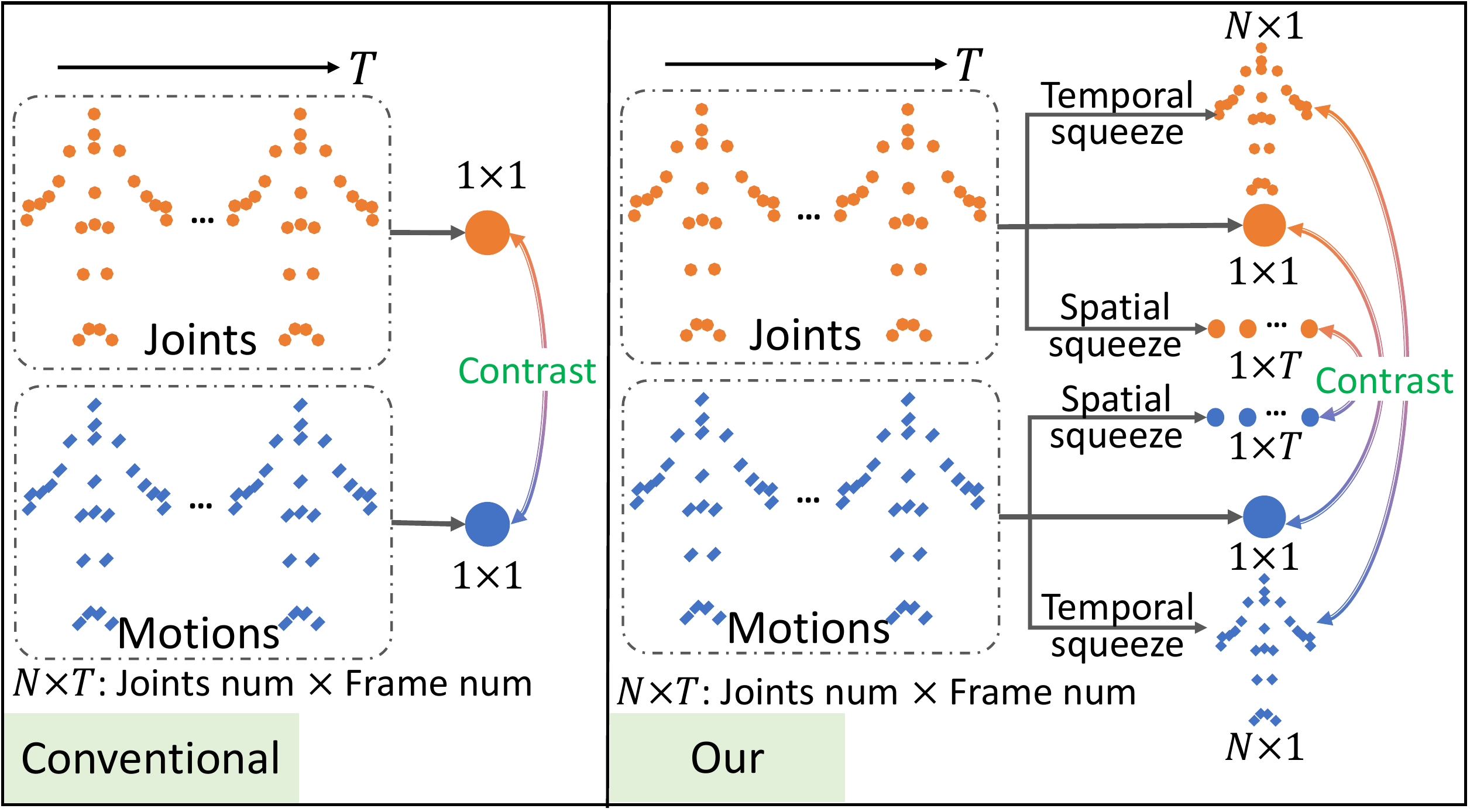}
    \caption{
    The main idea of this work. Conventional methods only contrast global features confusing the spatial- and temporal-specific information. Our method jointly contrasts temporal-squeezing features, spatial-squeezing features, and global features.
    }
    \label{idea}
\end{figure}

Currently, skeleton-based action recognition has achieved promising progress. In general, most existing skeleton-based methods can be grouped into four categories, namely Recurrent Neural Network (RNN)-based method, Convolutional Neural Network (CNN)-based method, Graph Convolutional Network (GCN)-based method, Transformer-based method~\cite{du2015hierarchical,liu2017skeleton,li2017skeleton,yan2018spatial,li2019actional,shi2019two,shi2020decoupled,wang2021iip}. 
Overall, most of the well-performing skeleton-based methods are trained in supervised learning, which relies on large amounts of labeled data. However, annotating skeletal data is always labor- and time-consuming. This prompts the development of techniques requiring only few or no labels, such as semi-supervised skeleton-based action recognition.

\begin{figure*}[!t]
    \centering
    \includegraphics[width=7.1in]{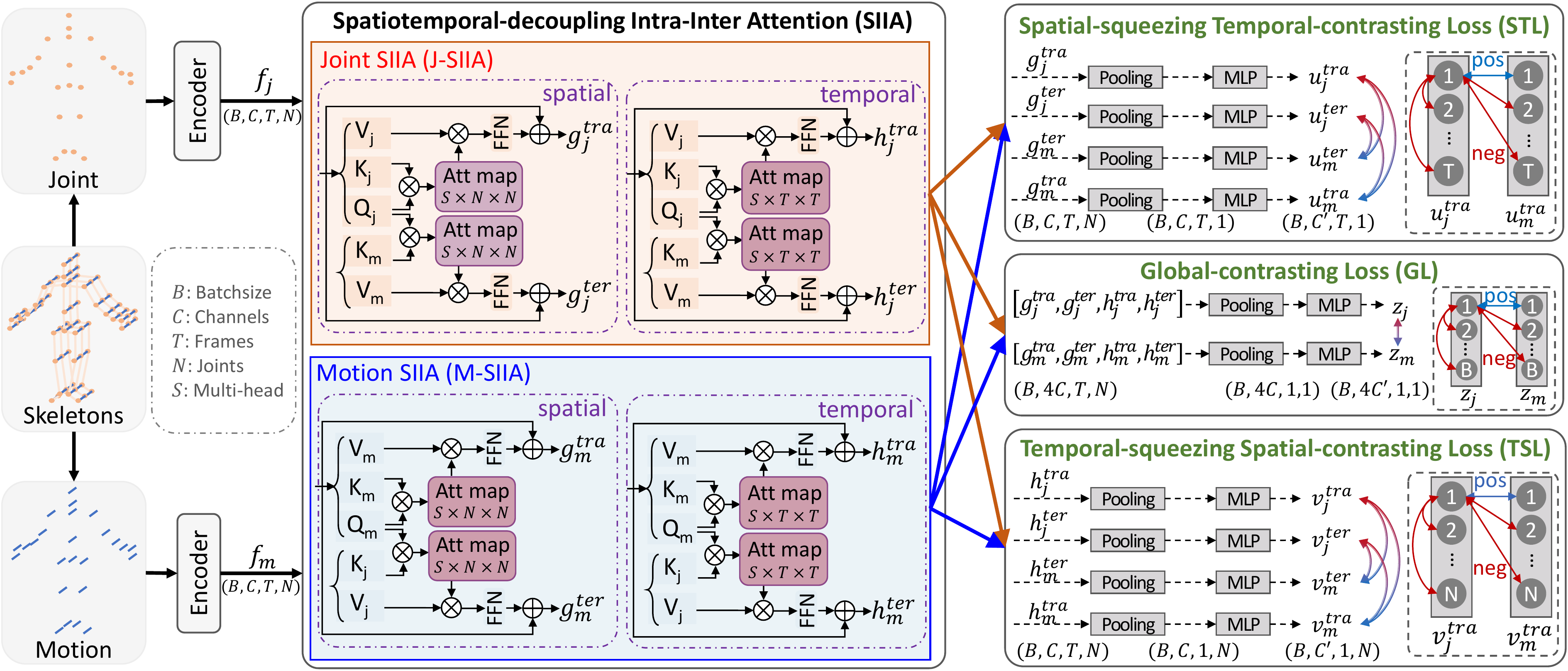}
    \caption{
    The overall framework of Spatiotemporal Decouple-and-Squeeze Contrastive Learning (SDS-CL). It is mainly composed of Encoder, Spatiotemporal-decoupling Intra-Inter Attention (SIIA) including Joint SIIA (J-SIIA) and Motion SIIA (M-SIIA), Temporal-squeezing Spatial-contrasting Loss (TSL), Spatial-squeezing Temporal-contrasting Loss (STL), and Global-contrasting Loss (GL). The joint and motion skeleton data are input into Encoder and SIIA to obtain the spatiotemporal-decoupling attentive features $\{g_*^{tra},g_*^{ter},h_*^{tra},h_*^{ter}\} (*=j,m)$ by calculating spatial- and temporal-decoupling intra-attention maps among joint/motion features, as well as spatial- and temporal-decoupling inter-attention maps between joint and motion features. Finally, STL, TSL, and GL aim to correspondingly contrast the spatial-squeezing features $\{u_*^{tra},u_*^{ter}\}$ at the frame level, temporal-squeezing features $\{v_*^{tra},v_*^{ter}\}$ at the joint level, and global features $\{z_j,z_m\}$ at the skeleton level for jointly training the whole framework. More details of SIIA can be found in Figure~\ref{attention}.
    }
    \label{framework}
\end{figure*}

Meanwhile, contrastive learning is applied to semi-supervised and unsupervised skeleton-based action recognition and has achieved impressive performance~\cite{li20213d,thoker2021skeleton}. Generally, most contrastive learning-based methods only contrast global
features mixing spatiotemporal information, which confuses the spatial- and temporal-specific information reflecting different semantic at the frame level and joint level. For example, for the action ``rub two hands together" and the action ``put the palms together", they have more spatial differences in the joint-level spatial dimension, so spatial-specific information is more important to distinguish them. But for action ``sitting down" and action ``standing up", they have more differences in the frame-level temporal dimension, so temporal-specific information is more important. 

Based on the above analysis, contrasting spatial- and temporal-specific features via contrastive learning is also beneficial to the skeleton-based action recognition task. Therefore, we propose a novel Spatiotemporal Decouple-and-Squeeze Contrastive Learning (SDS-CL) framework that jointly contrasts spatial-squeezing features, temporal-squeezing features, and global features for learning more abundant action representations. In SDS-CL framework, we design a new Spatiotemporal-decoupling Intra-Inter Attention
(SIIA), which includes Joint SIIA (J-SIIA) and Motion SIIA
(M-SIIA), to obtain the spatiotemporal-decoupling attentive features
for capturing spatiotemporal specific information. Specifically, we leverage J-SIIA and M-SIIA to calculate spatial- and
temporal-decoupling intra-attention maps among joint/motion
features, as well as spatial- and temporal-decoupling inter-attention maps between joint and motion features, respectively. Then, we present a novel Spatial-squeezing Temporal-contrasting Loss (STL), a novel Temporal-squeezing Spatial-contrasting Loss (TSL), and the Global-contrasting Loss (GL) to jointly measure the agreement between spatial-squeezing joint and motion features at the frame level, the agreement between temporal-squeezing joint and motion features at the joint level, as well as the agreement between global joint and motion features at the skeleton level.


The whole framework of the proposed SDS-CL is shown in Figure~\ref{framework}. SDS-CL mainly consists of Encoder~\cite{shi2020decoupled}, Spatiotemporal-decoupling Intra-Inter Attention (SIIA) including Joint SIIA (J-SIIA) and Motion SIIA (M-SIIA), Spatial-squeezing Temporal-contrasting Loss (STL), Temporal-squeezing Spatial-contrasting Loss (TSL), and Global-contrasting Loss (GL). First, we feed the joint and motion data generated from the raw skeletal data into Encoder to obtain joint features and motion features, respectively. Second, joint features and motion features are fed into J-SIIA and M-SIIA to obtain spatiotemporal-decoupling attentive features of joints and motions, followed by STL, TSL, and GL. In STL, TSL, and GL, we transform all spatiotemporal-decoupling attentive features into the spatial-squeezing features, temporal-squeezing features, global features, and then measure the distance between the spatial-squeezing/temporal-squeezing/global features of joints and motions. Here, STL, TSL, and GL are used together for training the whole SDS-CL framework in the pre-training phase, and in the finetune phase, some labeled data are fed to the pre-trained Encoder and a recognition head (including the MLP and softmax) in turn, where the latter is used for fine-tuning the former. Finally, we conduct experiments on four public datasets to validate the recognition performance of the proposed SDS-CL.


Overall, the main contributions in this work can be summarized as follows:
\begin{itemize}
		\item  To address the problem of semi-supervised skeleton-based action recognition, we propose a novel Spatiotemporal Decouple-and-Squeeze Contrastive Learning (SDS-CL) framework to comprehensively learn more abundant representations of skeleton-based actions by jointly contrasting spatial-squeezing features, temporal-squeezing features, and global features.
		\item We design a new Spatiotemporal-decoupling Intra-Inter Attention (SIIA) mechanism to capture more spatial- and temporal-specific information by calculating spatial- and temporal-decoupling intra-attention maps among joint/motion features, as well as spatial- and temporal-decoupling inter-attention maps between joint and motion features.
		\item We present a new Spatial-squeezing Temporal-contrasting Loss (STL), a new Temporal-squeezing Spatial-contrasting Loss (TSL), and the Global-contrasting Loss (GL) to correspondingly contrast spatial-squeezing features at frame level, temporal-squeezing features at joint level, and global features at skeleton level.
	\end{itemize}

The rest of this paper is organized as follows: Section~\ref{related work} describes related works including supervised, semi-supervised, unsupervised and self-supervised skeleton-based action recognition, contrastive learning, as well as self-attention mechanism. Section~\ref{methodology} presents the proposed SDS-CL in detail for semi-supervised skeleton-based action recognition. Results and analysis of extensive experiments are reported in Section~\ref{experiments}, and the conclusion is given in Section~\ref{conclusion}.

\section{Related Work}
\label{related work}
We briefly survey some works related to supervised, semi-supervised, unsupervised, and self-supervised skeleton-based action recognition. Besides, we also introduce contrastive learning and self-attention mechanism.
\subsection{Supervised Skeleton-based Action Recognition}
Generally, for the task of supervised skeleton-based action recognition, deep learning-based methods outperform traditional handcrafted-based methods~\cite{hussein2013human,vemulapalli2014human,liu2020adversarial,li2021memory,du2015hierarchical,liu2016spatio,shahroudy2016ntu,du2015skeleton,kim2017interpretable,liu2017enhanced,yan2018spatial,shi2019two,song2020stronger,chen2021channel,shi2020decoupled,wang2021iip,shi2022multiscale}. 

Specifically, RNN-based methods usually model skeleton data as a sequence of joints under a certain traversal way to learn the temporal dynamics between frames. For example, Du et al.~\cite{du2015hierarchical} considered utilizing RNN to model the hierarchical fusion of the five parts of the skeleton in the temporal space. CNN-based methods try to transform the skeleton data into the skeleton image for learning spatial representations, which is similar to image classification. For example, Du et al.~\cite{du2015skeleton} represented the skeleton sequences as the matrices by treating the joint coordinates as channels, which are fed into CNN for feature extraction. Compared with RNN-based methods and CNN-based methods, GCN-based methods can effectively explore the interdependence between skeleton joints by pre-designing the topology of skeleton data. For example, Yan et al.~\cite{yan2018spatial} designed a spatial and temporal graph convolutional network to model the skeletons in the graph structure manner, which regards joints and bones as nodes and edges of the skeleton graph. Since the design of topology in GCN is important for capturing the relationship of features, Chen et al.~\cite{chen2021channel} proposed different channel-wise topologies and aggregated joint features in different channels by GCN. Although GCN-based method has shown remarkable performance, its dependence on topological structure limits the modeling flexibility. In contrast to the limitations of the GCN-based method, transformer-based methods composed of self-attention blocks can learn more flexible relationship among all joints without any pre-designed topology. For example, Shi et al.~\cite{shi2020decoupled} leveraged self-attention block to capture the spatiotemporal dependencies among joints without any pre-defined connections of joints. To alleviate the noise effect brought by the individual joint, Wang et al.~\cite{wang2021iip} used a transformer to effectively encoder inter-part and intra-part dependencies. Overall, the above deep learning-based methods are trained in a supervised manner, which generally requires a larger number of labeled training data.

\subsection{Semi-Supervised Skeleton-based Action Recognition}
Semi-supervised learning indicates learning representations of data from both unlabeled data and labeled data, where the number of unlabeled data is generally larger than that of labeled data~\cite{chapelle2009semi}. For the semi-supervised skeleton-based action recognition task, how to learn effectively motion representations of unlabeled skeleton data is crucial~\cite{liu2020semi,si2020adversarial,li2020sparse,li2020iterate,tu2022joint}. Si et al.~\cite{si2020adversarial} presented a semi-supervised learning scheme to capture more discrimination representations by utilizing an adversarial regularization to align features from labeled and unlabeled data. To amend the limitation of needing to know all classes and learn more robust representations, Li et al.~\cite{li2020sparse} used an Encoder-Decoder RNN to learn the latent representations of unlabeled data, and then built an active learning framework to select labeled data based on cluster and classification uncertainty. Compared with prior semi-supervised methods of only considering motion information from joints, Tu et al.~\cite{tu2022joint} explored the motion transmission between joints and bones via encoder of GCN and decoder of a pose prediction head in a semi-supervised learning way. Overall, compared with supervised skeleton-based action recognition, many challenging problems target to the semi-supervised skeleton-based action recognition task are remaining. To address these problems, we attempt to leverage advanced contrastive learning and superior encoder mechanism of the transformer to learn motion representations from joints and motions on unlabeled data.

\subsection{Unsupervised and Self-Supervised Skeleton-based Action Recognition}
Unsupervised learning indicates learning representations of data from only unlabeled data, and self-supervised learning can be seen as one specific version of unsupervised learning. There are some studies based on encoder-decoder framework and contrastive learning framework for unsupervised and self-supervised skeleton-based action recognition~\cite{holden2015learning,zheng2018unsupervised,kundu2019unsupervised,lin2020ms2l,su2020predict,xu2020prototypical,rao2021augmented,gao2021contrastive,su2021self,yang2021skeleton,li20213d,thoker2021skeleton,tanfous2022and,liu2021imigue}. For encoder-decoder-based methods, to learn the long-term global motion dynamics, Zheng et al.~\cite{zheng2018unsupervised} introduced a conditional encoder-decoder framework with additional adversarial training tactics. To learn a separable feature representation, Su et al.~\cite{su2020predict} presented an encoder-decoder RNN to self-organize the hidden states into a feature space. 

Currently, some contrastive learning-based methods have achieved good performance on unsupervised or self-supervised skeleton-based action recognition tasks~\cite{rao2021augmented,gao2021contrastive,su2021self,guo2022contrastive,xu2020prototypical,lin2020ms2l,li20213d}. Specifically, some approaches investigate different augmentation strategies. For example, Rao et al.~\cite{rao2021augmented} exploited different augmentations to learn inherent action pattern representations in the contrastive action learning paradigm. Gao et al.~\cite{gao2021contrastive} designed the augmentations of compositions of viewpoints and distances to learn motion semantic invariance of distance and viewpoint. Su et al.~\cite{su2021self} constructed the speed-changed and motion-broken augmentation to learn the intrinsic dynamic motion consistency information. Guo et al.~\cite{guo2022contrastive} presented a distributional divergence minimization between the normal-augmented data and extreme-augmented data by bringing in extreme augmentation and normal augmentation.
Besides, some approaches explore various pretext tasks. For example, Xu et al.~\cite{xu2020prototypical} presented the pretext task including reverse prediction and prototypical contrast to learn low-level, high-level information and implicit semantic similarity. Lin et al.~\cite{lin2020ms2l} integrated multiple pretext tasks containing motion prediction, jigsaw puzzle recognition, and contrastive learning to learn more general action representations. Li et al.~\cite{li20213d} introduced the cross-view contrastive pretext task including joint view and motion view to learn more accurate representation with high-confidence positive/negative samples.

Overall, the above contrastive learning-based methods only contrast global features mixing spatiotemporal information, which confuses the spatial- and temporal-specific information. In this work, SDS-CL presents a new spatial-squeezing temporal-contrasting pretext task, a new temporal-squeezing spatial-contrasting pretext task, and an existing global-contrasting pretext task to additionally capture the spatial- and temporal-specific information from unlabeled data, except the global information.

\subsection{Contrastive Learning}
Contrastive learning based on instance discrimination has attracted wide attention in the field of representation learning~\cite{wu2018unsupervised,misra2020self,henaff2020data,chen2020improved,he2020momentum,tian2020contrastive,chen2020simple,caron2020unsupervised,grill2020bootstrap}. Among them, some methods adopted a memory bank to store the representation vectors for contrasting~\cite{wu2018unsupervised,misra2020self,he2020momentum,chen2020improved,tian2020contrastive}. For example, Wu et al.~\cite{wu2018unsupervised} maintained a memory bank for storing representations. He et al.~\cite{he2020momentum} built a dynamic dictionary with a momentum update based on a memory bank to keep the stored representations consistent. Some methods took negative samples from the within mini-batches rather than from a memory bank~\cite{henaff2020data,chen2020simple,caron2020unsupervised}. For example, Chen et al~\cite{chen2020simple} introduced the composition of multiple data augmentation operations and added a learnable nonlinear layer for improving the representation quality via contrasting learning in a within mini-batches. To relieve the computationally challenge with a large number of negative samples, Caron et al.~\cite{caron2020unsupervised} presented an in-batches contrast at the cluster level instead of the sample level. Inspired by these impressive contrastive losses, we presented a group of contrastive losses that thoroughly contrast features at different levels, i.e., spatial-squeezing temporal-contrasting loss at the frame level, and temporal-squeezing spatial-contrasting loss at the joint level, and global-contrasting loss at the skeleton level.

\subsection{Self-Attention Mechanism}
Self-attention mechanism aims to capture inter-relationship between each element of sequences by global context information~\cite{vaswani2017attention,khan2021transformers}. Specifically, the input sequences firstly are projected to queries, keys, and values by three learnable weight matrices, and then packed into matrices form (denoted by $Q$, $K$ of dimension $d_k$, $V$). Second, the dot product results of the query and all keys are multiplied by the scaling factor $\frac{1}{\sqrt{d_k}}$, and then normalized as the weights of values by a softmax operator. Finally, the output is the weighted sum of the values. A traditional self-attention formula is expressed as follows,
\begin{equation}
    Attention(Q,K,V) = softmax(\frac{QK^\top}{\sqrt{d_k}})V
\label{tra_attention}
\end{equation}
As the main spotlight in Transformer, self-attention plays an important role in Transformer for automatically capturing the receptive field. More details about self-attention can be found in some survey works of Transformer~\cite{han2022survey,khan2021transformers,lin2021survey,selva2022video}.
Based on traditional self-attention, we design a new Spatiotemporal-decoupling Intra-Inter Attention (SIIA), which additionally brings in inter-attention for capturing the interaction between joint and motion modalities, except the intra-attention (i.e., self-attention) for capturing the self-interaction among the single modality.
\section{Methodology}
\label{methodology}
\subsection{Overview of SDS-CL}
In this work, given a skeleton sequence as the input, we aim to learn a more abundant action representation in a semi-supervised way, for the downstream task, i.e., skeleton-based action recognition. Figure~\ref{framework} shows the overall framework of the proposed SDS-CL.
First, joint features $f_j$ and motion features $f_m$ are obtained by inputting joint data $\mathcal{X}_j$ and motion data $\mathcal{X}_m$ into Encoder, where $f_j, f_m \in \mathbb{R}^{ B\times C\times T\times N}$, $B$, $C$, $T$, and $N$ denote batch size, the number of channels, frames, and joints, respectively. 
Second, the joint features $f_j$ and motion features $f_m$ are fed into SIIA. We calculate two kinds of attention maps and obtain spatiotemporal-decoupling attentive features $\{g_*^{tra},g_*^{ter},h_*^{tra},h_*^{ter}\} (*=j,m)$, where $g_*^{tra}$, $g_*^{ter}$, $h_*^{tra}$, $h_*^{ter}$ $\in \mathbb{R}^{ B\times C\times T\times N}$. Here, the spatial- and temporal-decoupling intra-attention maps are calculated among joint/motion features. The spatial- and temporal-decoupling inter-attention maps are calculated between joint and motion features. Third, all spatiotemporal-decoupling attentive features are further fed into STL, GL, and TSL for contrastive learning. In STL, TSL, and GL, we obtain spatial-squeezing features $\{u^{tra}_*, u^{ter}_*\}$, temporal-squeezing features $\{v^{tra}_*, v^{ter}_*\}$, and global features $z_*$, followed by contrasting these features between joints and motions, where $u^{tra}_*$, $u^{ter}_* \in \mathbb{R}^{B\times C'\times T\times 1}$, $v^{tra}_*$, $v^{ter}_* \in \mathbb{R}^{B\times C'\times 1\times N}$, $z_* \in \mathbb{R}^{B\times 4C'\times 1\times 1}$. $\{u^{tra}_*, u^{ter}_*\}$ are obtained from $\{g^{tra}_*, g^{ter}_*\}$ by Pooling and MLP, followed by contrasting $\{u^{tra}_j, u^{ter}_j\}$ and $\{u^{tra}_m, u^{ter}_m\}$ at the frame level. $\{v^{tra}_*, v^{ter}_*\}$ are obtained from $\{h^{tra}_*, h^{ter}_*\}$ by Pooling and MLP, followed by contrasting $\{v^{tra}_j, v^{ter}_j\}$ and $\{v^{tra}_m, v^{ter}_m\}$ at the joint level. $z_*$ are obtained from $\{g^{tra}_*, g^{ter}_*, h^{tra}_*, h^{ter}_*\}$ by concatenating, Pooling and MLP, followed by contrasting $z_j$ and $z_m$ at the skeleton level. For convenience, some important notations are defined in Table~\ref{notations}.
\begin{table}[!t]
    \renewcommand{\arraystretch}{1.3}
    \centering
    \caption{Some notations and definitions.}
    \begin{tabular}{l|l}
        \hline
        Notation & Definition \\ \hline
        $f_j$ & Joint features\\
        $f_m$ & Motion features\\ 
        $g^{tra}_j$ & Spatial-decoupling intra-attentive features of joints\\
        $g^{tra}_m$ & Spatial-decoupling intra-attentive features of motions\\
        $g^{ter}_j$ & Spatial-decoupling inter-attentive features of joints\\
        $g^{ter}_m$ & Spatial-decoupling inter-attentive features of motions\\
        $h^{tra}_j$ & Temporal-decoupling intra-attentive features of joints\\
        $h^{tra}_m$ & Temporal-decoupling intra-attentive features of motions\\
        $h^{ter}_j$ & Temporal-decoupling inter-attentive features of joints\\
        $h^{ter}_m$ & Temporal-decoupling inter-attentive features of motions\\
        $u^{tra}_j$ & Spatial-squeezing features from $g^{tra}_j$\\
        $u^{tra}_m$ & Spatial-squeezing features from $g^{tra}_m$\\
        $u^{ter}_j$ & Spatial-squeezing features from $g^{ter}_j$\\
        $u^{ter}_m$ & Spatial-squeezing features from $g^{ter}_m$\\
        $v^{tra}_j$ & Temporal-squeezing features from $h^{tra}_j$\\
        $v^{tra}_m$ & Temporal-squeezing features from $h^{tra}_m$\\
        $v^{ter}_j$ & Temporal-squeezing features from $h^{ter}_j$\\
        $v^{ter}_m$ & Temporal-squeezing features from $h^{ter}_m$\\
        $z_j$ & Global features of joints\\
        $z_m$ & Global features of motions\\
        $\mathcal{L}^{STL}_1$ & STL among $\mathcal{U}_1 = u^{tra}_j \cup u^{tra}_m$ \\
        $\mathcal{L}^{STL}_2$ & STL among $\mathcal{U}_2 = u^{tra}_j \cup u^{ter}_m$ \\
        $\mathcal{L}^{STL}_3$ & STL among $\mathcal{U}_3 = u^{ter}_j \cup u^{tra}_m$ \\
        $\mathcal{L}^{STL}_4$ & STL among $\mathcal{U}_4 = u^{ter}_j \cup u^{ter}_m$ \\
        $\mathcal{L}^{TSL}_1$ & TSL among $\mathcal{V}_1 = v^{tra}_j \cup v^{tra}_m$ \\
        $\mathcal{L}^{TSL}_2$ & TSL among $\mathcal{V}_2 = v^{tra}_j \cup v^{ter}_m$ \\
        $\mathcal{L}^{TSL}_3$ & TSL among $\mathcal{V}_3 = v^{ter}_j \cup v^{tra}_m$ \\
        $\mathcal{L}^{TSL}_4$ & TSL among $\mathcal{V}_4 = v^{ter}_j \cup v^{ter}_m$ \\
        \hline
    \end{tabular}
    \label{notations}
\end{table}


\subsection{Spatiotemporal-decoupling Intra-Inter Attention (SIIA)}
Spatiotemporal-decoupling Intra-Inter Attention (SIIA) includes Joint SIIA (J-SIIA) and Motion SIIA (M-SIIA) for decoupling and capturing spatiotemporal specific information, as well as assisting in the construction of corresponding spatial/temporal-squeezing features for the subsequent contrast.

\subsubsection{Joint SIIA}
\begin{figure}[!t]
    \centering
    \includegraphics[width=3.4in]{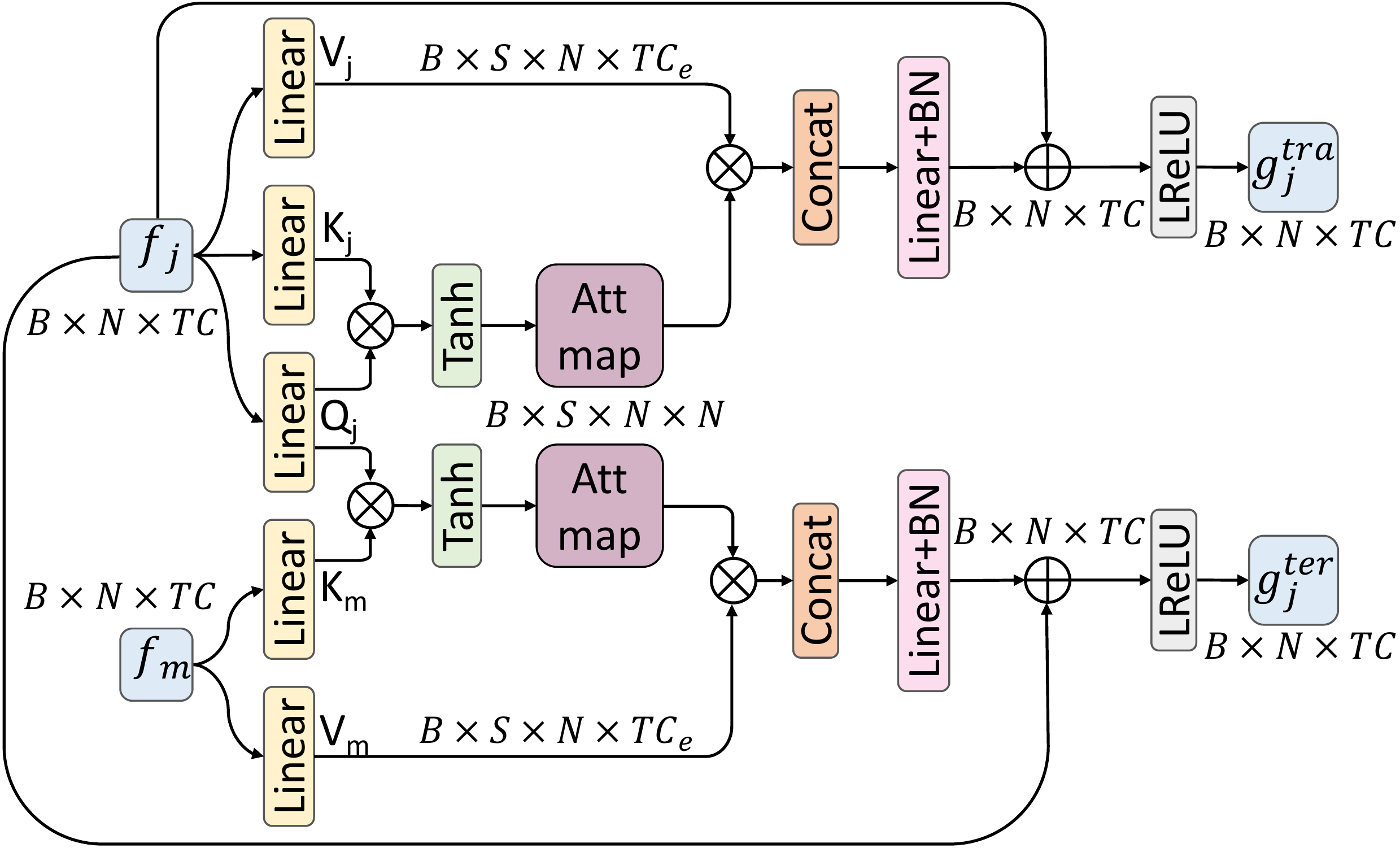}
    \caption{Architecture of spatial part in J-SIIA. $f_j$ and $f_m$ are joint features and motion features, respectively. Linear is the linear projection layer. $\otimes$ denotes the matrix multiplication. BN is the batch norm. $\oplus$ denotes the residual connection addition. LReLU is the leaky ReLU activation function. $g^{tra}_j$ and $g^{ter}_j$ are spatial-decoupling attentive features in joint modality.}
    \label{attention}
\end{figure}

In J-SIIA, we project joint features $f_j$ to $Q_j$ as the query vector, and then its corresponding intra-key vector $K_j$ and intra-value vector $V_j$ can be defined by projecting $f_j$, its corresponding inter-key vector $K_m$ and inter-value vector $V_m$ can be obtained by projecting $f_m$. Here, the dimensions of $\{Q_j$, $K_j$, $V_j$, $K_m$, $V_m\}$ are $\mathbb{R}^{B\times S\times N\times TC_e}$ and $\mathbb{R}^{B\times S\times T\times NC_e}$ ($S\times C_e=C$) for spatial-decoupling and temporal-decoupling, respectively.
Formally, the spatial-decoupling intra-attention map $\mathcal{A}_{tra}^{spa}$ and spatial-decoupling inter-attention map $\mathcal{A}_{ter}^{spa}$ can be computed as follows,
\begin{equation}
    \begin{cases}
        \displaystyle\mathcal{A}_{tra}^{spa} = \tanh(\frac{Q_j(K_j)^\top}{\sqrt{\hat{C}^{spa}}}), \; Q_j,K_j\in \mathbb{R}^{{\color{black}B\times} S\times N\times TC_e}; \\
        \displaystyle\mathcal{A}_{ter}^{spa} = \tanh(\frac{Q_j(K_m)^\top}{\sqrt{\hat{C}^{spa}}}), \; Q_j,K_m\in \mathbb{R}^{{\color{black}B\times} S\times N\times TC_e}
    \end{cases}
    \label{eq_spatial_map}
\end{equation}
where $\mathcal{A}_{tra}^{spa}$, $\mathcal{A}_{ter}^{spa} \in \mathbb{R}^{{\color{black}B\times} S\times N\times N}$, $\hat{C}^{spa} = C_e\times T$, $\top$ is the transpose operation of matrix/vector, and $S$ is the number of multi-head. Different from the activation function softmax used in traditional self-attention of Eq.~\eqref{tra_attention}, the results produced by tanh contain negative values and are not limited to positive values, thus tanh is more flexible than softmax for calculating attention maps. So we use tanh instead of softmax as the activation function~\cite{shi2020decoupled, qiu2022spatio, shi2021star}. In Eq.~\eqref{eq_spatial_map}, $\mathcal{A}_{tra}^{spa}$ and $\mathcal{A}_{ter}^{spa}$ capture the spatial-specific relationship to prompt the learned representation to pay more attention to the key intra-modal spatial information among joints, and the key inter-modal spatial information between joints and motions, respectively. 
Similarly, the temporal-decoupling intra-attention map $\mathcal{A}_{tra}^{tem}$ and temporal-decoupling inter-attention map $\mathcal{A}_{ter}^{tem}$ can be computed as follows,
\begin{equation}
    \begin{cases}
        \displaystyle\mathcal{A}_{tra}^{tem} = \tanh(\frac{Q_j(K_j)^\top}{\sqrt{\hat{C}^{tem}}}), \; Q_j,K_j\in \mathbb{R}^{{\color{black}B\times} S\times T\times NC_e}; \\
        \displaystyle\mathcal{A}_{ter}^{tem} = \tanh(\frac{Q_j(K_m)^\top}{\sqrt{\hat{C}^{tem}}}), \; Q_j,K_m\in \mathbb{R}^{{\color{black}B\times} S\times T\times NC_e}
    \end{cases}
    \label{eq_temporal_map}
\end{equation}
where $\mathcal{A}_{tra}^{tem}$, $\mathcal{A}_{ter}^{tem} \in \mathbb{R}^{{\color{black}B\times} S\times T\times T}$, and $\hat{C}^{tem} = C_e\times N$. $\mathcal{A}_{tra}^{tem}$ and $\mathcal{A}_{ter}^{tem}$ capture the temporal-specific relationship to prompt the learned representation to pay more attention to the key intra-modal temporal information among joints, and the key inter-modal temporal information between joints and motions, respectively.
Finally, the spatiotemporal-decoupling attentive features $\{g_j^{tra},g_j^{ter},h_j^{tra},h_j^{ter}\}$ in joint modality are calculated as follows,
\begin{equation}
\label{eq3}
    g_j^{tra} = \sigma(\phi(Concat(\mathcal{A}_{tra}^{spa}V_j))+f_j)
\end{equation}
\begin{equation}
\label{eq4}
    g_j^{ter} = \sigma(\phi(Concat(\mathcal{A}_{ter}^{spa}V_m))+f_j)
\end{equation}
\begin{equation}
\label{eq5}
    h_j^{tra} = \sigma(\phi(Concat(\mathcal{A}_{tra}^{tem}V_j))+f_j)
\end{equation}
\begin{equation}
\label{eq6}
    h_j^{ter} = \sigma(\phi(Concat(\mathcal{A}_{ter}^{tem}V_m))+f_j)
\end{equation}
where $\phi$ denotes Feed Forward Network (FFN) including a linear layer and a batch norm layer, $\sigma$ denotes the leaky ReLU activation function, and $Concat$ denotes the concatenation of multiple result vectors. For example, spatial-decoupling intra-attentive feature $g_j^{tra}$ comes from $\mathcal{A}_{tra}^{spa}$ multiplied by $V_j$ in joint modality, and spatial-decoupling inter-attentive feature $g_j^{ter}$ comes from $\mathcal{A}_{ter}^{spa}$ multiplied by $V_m$ in motion modality. In J-SIIA, the detailed architecture of its spatial part is shown in Figure~\ref{attention}.

\subsubsection{Motion SIIA}

In M-SIIA, we also project motion features and joint features to obtain the spatial- and temporal-decoupling intra-attention/inter-attention maps $\mathcal{B}_{tra}^{spa}$, $\mathcal{B}_{ter}^{spa}$, $\mathcal{B}_{tra}^{tem}$, $\mathcal{B}_{ter}^{tem}$, similar to Eq.~\eqref{eq_spatial_map},~\eqref{eq_temporal_map}, which are expressed as follows,
\begin{equation}
    \begin{cases}
        \displaystyle\mathcal{B}_{tra}^{spa} = \tanh(\frac{Q_m(K_m)^\top}{\sqrt{\hat{C}^{spa}}}), \; Q_m,K_m\in \mathbb{R}^{B\times S\times N\times TC_e}; \\
        \displaystyle\mathcal{B}_{ter}^{spa} = \tanh(\frac{Q_m(K_j)^\top}{\sqrt{\hat{C}^{spa}}}), \; Q_m,K_j\in \mathbb{R}^{B\times S\times N\times TC_e}
    \end{cases}
    \label{eq_motion_spatial}
\end{equation}
\begin{equation}
    \begin{cases}
        \displaystyle\mathcal{B}_{tra}^{tem} = \tanh(\frac{Q_m(K_m)^\top}{\sqrt{\hat{C}^{tem}}}), \; Q_m,K_m\in \mathbb{R}^{B\times S\times T\times NC_e}; \\
        \displaystyle\mathcal{B}_{ter}^{tem} = \tanh(\frac{Q_m(K_j)^\top}{\sqrt{\hat{C}^{tem}}}), \; Q_m,K_j\in \mathbb{R}^{B\times S\times T\times NC_e}
    \end{cases}
    \label{eq_motion_temporal}
\end{equation}
$\mathcal{B}_{tra}^{spa}$ and $\mathcal{B}_{tra}^{tem}$ capture the key intra-modal spatial/temporal information among motions. $\mathcal{B}_{ter}^{spa}$ and $\mathcal{B}_{ter}^{tem}$ capture the key inter-modal spatial/temporal information between motions and joints. Lastly, the spatiotemporal-decoupling attentive features $\{g_m^{tra},g_m^{ter},h_m^{tra},h_m^{ter}\}$ in motion modality, similar to Eq.~\eqref{eq3}-\eqref{eq6}, which are expressed as follows,
\begin{equation}
\label{eq7}
    g_m^{tra} = \sigma(\phi(Concat(\mathcal{B}_{tra}^{spa}V_m))+f_m)
\end{equation}
\begin{equation}
\label{eq8}
    g_m^{ter} = \sigma(\phi(Concat(\mathcal{B}_{ter}^{spa}V_j))+f_m)
\end{equation}
\begin{equation}
\label{eq9}
    h_m^{tra} = \sigma(\phi(Concat(\mathcal{B}_{tra}^{tem}V_m))+f_m)
\end{equation}
\begin{equation}
\label{eq10}
    h_m^{ter} = \sigma(\phi(Concat(\mathcal{B}_{ter}^{tem}V_j))+f_m)
\end{equation}

\subsection{Training Objective}
In our semi-supervised learning way, the whole framework of SDS-CL is first pre-trained on unlabeled data by the designed contrastive learning losses, and then the pre-trained Encoder and a recognition head are together fine-tuned on the labeled data. In the pre-training phase, the proposed SDS-CL is trained in an end-to-end strategy, and the overall contrastive loss $\mathcal{L}$ consists of Spatial-squeezing Temporal-contrasting Loss (STL) $\mathcal{L}^{STL}$, Temporal-squeezing Spatial-contrasting Loss (TSL) $\mathcal{L}^{TSL}$, and Global-contrasting Loss (GL) $\mathcal{L}^{GL}$, which is summarized as follows,
\begin{equation}
    \mathcal{L} = \mathcal{L}^{STL} + \mathcal{L}^{TSL} + \mathcal{L}^{GL}
\end{equation}
In the followings, we introduce these three types of losses in detail.
\subsubsection{Spatial-squeezing Temporal-contrasting Loss (STL)}
In order to promote the model to learn more temporal-specific information, Spatial-squeezing Temporal-contrasting Loss (STL) is presented to contrast spatial-squeezing features at the frame level. As shown in Figure~\ref{framework}, spatial-decoupling attentive features $\{g^{tra}_j, g^{ter}_j, g^{tra}_m, g^{ter}_m\}$ are obtained from SIIA. Then, $\{g^{tra}_j, g^{ter}_j, g^{tra}_m, g^{ter}_m\}$ are converted to spatial-squeezing features $\{u^{tra}_j, u^{ter}_j, u^{tra}_m, u^{ter}_m \in \mathbb{R}^{B\times C'\times T\times 1}\}$ by Pooling and MLP, followed by contrasting these spatial-squeezing features between joint and motion. Formally, $\mathcal{L}^{STL}_1$ contrasting between $u^{tra}_j$ and $u^{tra}_m$ is formulated on $\mathcal{U}_1 = u^{tra}_j \cup u^{tra}_m$ ($\mathcal{U}_1 \in \mathbb{R}^{B\times C'\times 2T\times 1}$) as follows,
\begin{equation}
    \mathcal{L}^{STL}_1 = -\frac{1}{2TB}\sum_{b=1}^B\sum_{i=1}^{2T}\log\frac{\exp(\langle\mu_i, \mu_{\tilde{i}}\rangle{\color{black}/\tau})}{\sum\nolimits_{k=1}^{2T}\mathds{1}_{[k\not=i]}\exp(\langle\mu_i,\mu_k\rangle/\tau)}
\end{equation}
where $\mu_i$, $\mu_{\tilde{i}}\in \mathcal{U}_1$, $\mu_i$ and $\mu_{\tilde{i}}$ are the representations of the same frame from joint and motion modalities, respectively, $\langle\mu_i,\mu_{\tilde{i}}\rangle=\mu_i^\top \mu_{\tilde{i}}/\| \mu_i\| \| \mu_{\tilde{i}}\|$, $\mathds{1}\in \{0,1\}$ is an indicator function that equals 1 if $k\not=i$, and $\tau$ is a temperature parameter. Meanwhile, the formula for $\mathcal{L}^{STL}_2$ contrasting between $u^{tra}_j$ and $u^{ter}_m$, $\mathcal{L}^{STL}_3$ contrasting between $u^{ter}_j$ and $u^{tra}_m$, $\mathcal{L}^{STL}_4$ contrasting between $u^{ter}_j$ and $u^{ter}_m$ is similar to that of $\mathcal{L}^{STL}_1$. Finally, STL is defined as follows,
\begin{equation}
    \mathcal{L}^{STL} = \mathcal{L}^{STL}_1 + \mathcal{L}^{STL}_2 + \mathcal{L}^{STL}_3 + \mathcal{L}^{STL}_4
\end{equation}
\subsubsection{Temporal-squeezing Spatial-contrasting Loss (TSL)}
To facilitate the model to learn more spatial-specific information, Temporal-squeezing Spatial-contrasting Loss (TSL) is designed to contrast temporal-squeezing features at the joint level. On the whole, the process of TSL is similar to that of STL. Specifically, as shown in Figure~\ref{framework}, temporal-decoupling attentive features $\{h^{tra}_j, h^{ter}_j, h^{tra}_m, h^{ter}_m\}$ are transformed to temporal-squeezing features $\{v^{tra}_j, v^{ter}_j, v^{tra}_m, v^{ter}_m\}$, which then are contrasted at the joint level by $\mathcal{L}^{TSL}=\mathcal{L}^{TSL}_1 + \mathcal{L}^{TSL}_2 + \mathcal{L}^{TSL}_3 + \mathcal{L}^{TSL}_4$. Here, $\mathcal{L}^{TSL}_1$ is defined as follows,
\begin{equation}
    \mathcal{L}^{TSL}_1 = -\frac{1}{2NB}\sum_{b=1}^B\sum_{i=1}^{2N}\log\frac{\exp(\langle\nu_i, \nu_{\tilde{i}}\rangle{\color{black}/\tau})}{\sum\nolimits_{k=1}^{2N}\mathds{1}_{[k\not=i]}\exp(\langle\nu_i,\nu_k\rangle/\tau)}
\end{equation}
where $\nu_i$, $\nu_{\tilde{i}}\in \mathcal{V}_1$, $\mathcal{V}_1 = v^{tra}_j \cup v^{tra}_m$ ($\mathcal{V}_1 \in \mathbb{R}^{B\times C'\times 1\times 2N}$), $\nu_i$ and $\nu_{\tilde{i}}$ are representations of the same joint from joint and motion modalities, respectively.
\subsubsection{Global-contrasting Loss (GL)}
Global-contrasting Loss (GL) is used for learning global information by contrasting global features between joints and motions at the skeleton level. As shown in Figure~\ref{framework}, spatiotemporal-decoupling attentive features $\{g^{tra}_j, g^{ter}_j, h^{tra}_j, h^{ter}_j\}$ and $\{g^{tra}_m, g^{ter}_m, h^{tra}_m, h^{ter}_m\}$ are respectively converted to global features $z_j$ and $z_m$ by concatenating, Pooling, and MLP. Then, $z_j$ and $z_m$ are contrasted via GL, which is defined as follows,
\begin{equation}
    \mathcal{L}^{GL} = -\frac{1}{2B}\sum_{i=1}^{2B}\log\frac{\exp(\langle\omega_i, \omega_{\tilde{i}}\rangle{\color{black}/\tau})}{\sum\nolimits_{k=1}^{2B}\mathds{1}_{[k\not=i]}\exp(\langle\omega_i,\omega_k\rangle/\tau)}
\end{equation}
where $\omega_i$, $\omega_{\tilde{i}} \in \Omega$, $\Omega = z_j \cup z_m$ ($\Omega \in \mathbb{R}^{2B\times 4C'\times 1\times 1}$), $\omega_i$ and $\omega_{\tilde{i}}$ are the representations of the same skeleton sequence from joint and motion modalities, respectively. Algorithm~1 summarizes the main implementations of SDS-CL.
 
 \begin{figure}[!t]
		\label{alg:SDS-CL}
		\renewcommand{\algorithmicrequire}{\textbf{Input:}}
        \removelatexerror
		\begin{algorithm}[H]
			\caption{Spatiotemporal Decouple-and-Squeeze Contrastive Learning (SDS-CL)}
			\begin{algorithmic}
			    \REQUIRE 
			    \STATE $\mathcal{X}_j$, $\mathcal{X}_m$ : joint data, and motion data
			    \STATE $K$ : total epochs
			    \STATE $\tau$ : temperature parameter
		        \FOR{$k=1$ \TO $K$}
		        \STATE $f_j$, $f_m$ = Encoder($\mathcal{X}_j$), Encoder($\mathcal{X}_m$)
		        \STATE
		        \STATE $//$ {\textit {Spatiotemporal-decoupling Intra-Inter Attention (SIIA)}}
		        \begin{equation}
		            \begin{aligned}
		                &Q_j, K_{j/m}, V_{j/m} = W^Q_jf_j, W^K_{j/m}f_{j/m}, W^V_{j/m}f_{j/m} \notag \\
		                &\mathcal{A}_{tra}^{spa} = \tanh(\frac{Q_j(K_j)^\top}{\sqrt{\hat{C}^{spa}}}) \quad\mathcal{A}_{tra}^{tem} = \tanh(\frac{Q_j(K_j)^\top}{\sqrt{\hat{C}^{tem}}}) \notag \\
		                &\mathcal{A}_{ter}^{spa} = \tanh(\frac{Q_j(K_m)^\top}{\sqrt{\hat{C}^{spa}}}) \quad\mathcal{A}_{ter}^{tem} = \tanh(\frac{Q_j(K_m)^\top}{\sqrt{\hat{C}^{tem}}}) \notag \\
		                &g_j^{tra} = \sigma(\phi({\text{Concat}}(\mathcal{A}_{tra}^{spa}V_j))+f_j) \notag \\
		                &g_j^{ter} = \sigma(\phi({\text{Concat}}(\mathcal{A}_{ter}^{spa}V_m))+f_j) \notag \\
		                &h_j^{tra} = \sigma(\phi({\text{Concat}}(\mathcal{A}_{tra}^{tem}V_j))+f_j) \notag \\
		                &h_j^{ter} = \sigma(\phi({\text{Concat}}(\mathcal{A}_{ter}^{tem}V_m))+f_j) \notag
		            \end{aligned}
                \end{equation}
                \STATE $\{g_m^{tra},g_m^{ter},h_m^{tra},h_m^{ter}\}$ similar to $\{g_j^{tra},g_j^{ter},h_j^{tra},h_j^{ter}\}$
                \STATE
                \STATE $//$ {\textit {Spatial-squeezing Temporal-contrasting Loss (STL)}}
                \begin{equation}
                    \begin{aligned}
                        &\mathcal{L}^{STL}_1 = -\frac{1}{2TB}\sum_{b=1}^B\sum_{i=1}^{2T}\log\frac{\exp(\langle\mu_i,\mu_{\tilde{i}}\rangle{\color{black}/\tau})}{\sum\nolimits_{k=1}^{2T}\mathds{1}_{[k\not=i]}\exp(\langle\mu_i,\mu_k\rangle/\tau)} \notag \\
                        &\mathcal{L}^{STL} = \mathcal{L}^{STL}_1 + \mathcal{L}^{STL}_2 + \mathcal{L}^{STL}_3 + \mathcal{L}^{STL}_4 \notag
                    \end{aligned}
                \end{equation}
                \STATE
                \STATE $//$ {\textit {Temporal-squeezing Spatial-contrasting Loss (TSL)}}
                \begin{equation}
                    \begin{aligned}
                        &\mathcal{L}^{TSL}_1 = -\frac{1}{2NB}\sum_{b=1}^B\sum_{i=1}^{2N}\log\frac{\exp(\langle\nu_i, \nu_{\tilde{i}}\rangle{\color{black}/\tau})}{\sum\nolimits_{k=1}^{2N}\mathds{1}_{[k\not=i]}\exp(\langle\nu_i,\nu_k\rangle/\tau)} \notag \\
                        &\mathcal{L}^{TSL} = \mathcal{L}^{TSL}_1 + \mathcal{L}^{TSL}_2 + \mathcal{L}^{TSL}_3 + \mathcal{L}^{TSL}_4 \notag
                    \end{aligned}
                \end{equation}
                \STATE
                \STATE $//$ {\textit {Global-contrsting Loss (GL)}}
                \begin{equation}
                    \begin{aligned}
                        &\mathcal{L}^{GL} = -\frac{1}{2B}\sum_{i=1}^{2B}\log\frac{\exp(\langle\omega_i, \omega_{\tilde{i}}\rangle{\color{black}/\tau})}{\sum\nolimits_{k=1}^{2B}\mathds{1}_{[k\not=i]}\exp(\langle\omega_i,\omega_k\rangle/\tau)} \notag
                    \end{aligned}
                \end{equation}
                \STATE
                \STATE $//$ {\textit {Objective function}}
                \STATE $\mathcal{L} = \mathcal{L}^{STL} + \mathcal{L}^{TSL} + \mathcal{L}^{GL}$
                \STATE Update all parameters using Stochastic Gradient Descent (SGD) with stop gradient strategy to minimize $\mathcal{L}$
		        \ENDFOR
			\end{algorithmic}
		\end{algorithm}
	\end{figure}
 
\section{Experiments}

In the semi-supervised skeleton-based action recognition
task, we conduct experiments on public datasets to evaluate
the performance of the proposed SDS-CL method compared
with some competitive methods.
\label{experiments}

\subsection{Dataset}
In the experiments, we evaluate the performance of the proposed SDS-CL on four public datasets, i.e., NTU RGB+D~\cite{shahroudy2016ntu}, Northwestern-UCLA~\cite{wang2014cross}, NTU RGB+D 120~\cite{liu2019ntu}, and Kinetics-Skeleton~\cite{kay2017kinetics}.

\textbf{NTU RGB+D dataset~\cite{shahroudy2016ntu}.} NTU RGB+D is a large-scale human action dataset shot by three Microsoft Kinetic v2 sensors for skeleton-based action recognition. It contains 56,578 samples covering 60 action categories that are performed by 40 different subjects ranging in age from 10 to 35, and each sample has up to two subjects consisting of 25 key joints. For experiments, it is recommended by the authors of this dataset to be divided into two benchmarks, i.e., Cross-Subject (CS) and Cross-View (CV). In CS, training data and validation data are divided by different subjects, where training data includes 40,091 samples from 20 subjects, and validation data includes 16,487 samples from the remaining 20 subjects. In CV, training data and validation data are divided by different sensor views, where training data contains 37,646 samples from sensor 2 and 3, and validation data contains 18,932 samples from sensor 1. For the task of semi-supervised skeleton-based action recognition, we follow the most popular setting where 5\%, 10\%, 20\%, and 40\% of labeled data are used for training, respectively.

\textbf{Northwestern-UCLA (NW-UCLA) dataset~\cite{wang2014cross}.} NW-UCLA contains 1,494 samples collected from 10 different subjects in 10 action classes by three Microsoft Kinect v1 sensors for skeleton-based action recognition, where each sample includes one subject composed of 20 joints. The evaluation benchmark is that training data contains 1,018 samples from sensor 1, 2, and validation data contains 476 samples from sensor 3. For the task of semi-supervised skeleton-based action recognition, we follow the most popular setting where 5\%, 15\%, 30\%, and 40\% of labeled data are used for training, respectively.

\textbf{NTU RGB+D 120 dataset~\cite{liu2019ntu}.} NTU RGB+D 120 extended from NTU RGB+D, as a larger-scale dataset, contains 113,945 samples of 120 human action classes that performed by 106 subjects. It has been defined two benchmarks well, i.e., Cross-Subject (CS) and Cross-Setup (CE). In CS, 63,026 training data and 50,919 testing data are collected by 53 different subjects, respectively. In CE, there are 32 different setup IDs, where 54,468/59,477 action sequences with even/odd setup IDs are used for training/testing. It is noted that, on the following protocol of pretrain + linear evaluation (details in Section~\ref{EE}), the training data includes the whole labeled data.

\textbf{Kinetics-Skeleton dataset~\cite{kay2017kinetics}.} Kinetics-Skeleton involves 260,000 2D skeleton sequences of 400 human action classes extracted by the OpenPose~\cite{cao2017realtime} toolbox, where 240,000 and 20,000 skeleton sequences are used for training and testing, respectively. It is noted that, on the following protocol of pretrain + finetune (details in Section~\ref{EE}), the training data includes the whole labeled data.

\begin{table*}[ht]
\renewcommand{\arraystretch}{1.3}
\caption{Recognition accuracies (\%) obtained by different methods on NTU RGB+D dataset (Cross-Subject (CS) and Cross-View (CV)) with 5\%, 10\%, 20\%, and 40\% labeled data of training set. The superscripts $^\ddagger$ and $^\dagger$ indicate the semi-supervised and unsupervised methods, respectively.}
\label{NTU}
\centering
\begin{tabular}{l||c|c||c|c||c|c||c|c}
\hline\hline
\multirow{2}{*}{Method}&
\multicolumn{2}{c||}{5\%} & \multicolumn{2}{c||}{10\%} & \multicolumn{2}{c||}{20\%} & \multicolumn{2}{c}{40\%} \\
\cline{2-9}
& CS & CV & CS & CV & CS & CV & CS & CV \\ \hline
$^\ddagger$S$^{4}$L \cite{zhai2019s4l} & 48.4  & 55.1  & 58.1  & 63.6 & 63.1  & 71.1  & 68.2  & 76.9 \\ \hline
$^\ddagger$Pseudolabels \cite{lee2013pseudo} & 50.9  & 56.3  & 58.4  & 65.8 & 63.9  & 71.2  & 69.5  & 77.7  \\ \hline
$^\ddagger$VAT \cite{miyato2018virtual} & 51.3  & 57.9  & 60.3  & 66.3 & 65.6  & 72.6  & 70.4  & 78.6 \\ \hline
$^\ddagger$VAT+EntMin \cite{grandvalet2005semi} & 51.7  & 58.3  & 61.4  & 67.5 & 65.9  & 73.3  & 70.8  & 78.9 \\ \hline
$^\ddagger$ASSL \cite{si2020adversarial} & 57.3  & 63.6  & 64.3  & 69.8 & 68.0  & 74.7  & 72.3  & 80.0 \\ \hline
$^\ddagger$AL+K \cite{li2020sparse} & 57.8 & - & 62.9 & - & - & - & - & - \\ \hline
$^\ddagger$CD-JBF-GCN \cite{tu2022joint} & 61.8 & 65.3 & 71.7 & 78.0 & 78.4 & 85.9 & 83.2 & 90.9 \\ \hline
$^\dagger$AS-CAL \cite{rao2021augmented} & - & - & 52.2 & 57.3 & - & - & - & - \\ \hline
$^\dagger$LongT GAN \cite{zheng2018unsupervised} & - & - & 62.0 & - & - & - & - & - \\ \hline
$^\dagger$Holden et al. \cite{holden2015learning} & - & - & - & - & - & - & 72.9 & 81.1 \\ \hline
$^\dagger$EnGAN-PoseRNN \cite{kundu2019unsupervised} & - & - & - & - & - & - & 78.7 & 86.5 \\ \hline
$^\dagger$MS$^{2}$L \cite{lin2020ms2l} & - & - & 65.2 & - & - & - & - & - \\ \hline
$^\dagger$Skeleton-Contrastive \cite{thoker2021skeleton} & 59.6 & 65.7 & 65.9 & 72.5 & 70.8 & 78.2 & - & - \\ \hline
$^\dagger$3s-Colorization \cite{yang2021skeleton} & 65.7 & 70.3 & 71.7 & 78.9 & 76.4 & 82.7 & 79.8 & 86.8 \\ \hline
$^\dagger$3s-CrosSCLR \cite{li20213d} & - & - & 74.4 & 77.8 & - & - & - & - \\ \hline
$^\dagger${\color{black}3s-AimCLR} \cite{guo2022contrastive} & - & - & \textbf{78.2} & 81.6 & - & - & - & - \\ \hline \hline
SDS-CL (Ours) & \textbf{71.3} & \textbf{75.3} & 77.2 & \textbf{83.0} & \textbf{82.2} & \textbf{86.4} & \textbf{85.7} & \textbf{91.1} \\ \hline
\end{tabular}
\end{table*}

\begin{table}[!t]
\renewcommand{\arraystretch}{1.3}
\caption{Recognition accuracies (\%) obtained by different methods on NW-UCLA with 5\%, 15\%, 30\%, and 40\% labeled data of training set. The superscripts $^\ddagger$ and $^\dagger$ indicate the semi-supervised and unsupervised methods, respectively.}
\label{NW-UCLA}
\centering
\begin{tabular}{l||c|c|c|c}
  \hline\hline
  Method & 5\% & 15\% & 30\% & 40\% \\ \hline
  $^\ddagger$S$^{4}$L \cite{zhai2019s4l} & 35.3 & 46.6 & 54.5 & 60.6 \\ \hline
  $^\ddagger$Pseudolabels \cite{lee2013pseudo} & 35.6 & 48.9 & 60.6 & 65.7 \\ \hline
  $^\ddagger$VAT \cite{miyato2018virtual} & 44.8 & 63.8 & 73.7 & 73.9 \\ \hline
  $^\ddagger$VAT+EntMin \cite{grandvalet2005semi} & 46.8 & 66.2 & 75.4 & 75.6 \\ \hline
  $^\ddagger$ASSL \cite{si2020adversarial} & 52.6 & 74.8 & 78.0 & 78.4 \\ \hline
  $^\ddagger$AL+K \cite{li2020sparse} & 63.6 & 76.8 & 77.2 & 78.9 \\ \hline
  $^\dagger$MS$^{2}$L \cite{lin2020ms2l} & - & 60.5 & - & - \\ \hline\hline
  SDS-CL (Ours) & \textbf{67.0} & \textbf{78.2} & \textbf{79.3} & \textbf{82.8} \\ \hline
\end{tabular}
\end{table}

\subsection{Experimental Setting and Implementation}
In the data pre-processing phase, all skeleton data from NTU RGB+D, NW-UCLA, NTU RGB+D 120, and Kinetics-Skeleton are both sampled to the length of 50 frames. In the semi-supervised setting of NTU RGB+D, training data includes around 33(5\%), 66(10\%), 132(20\%), 264(40\%) labeled data for each class under CS benchmark, and training data includes around 31(5\%), 62(10\%), 124(\%), 248(\%) labeled data for each class under CV benchmark. In the semi-supervised setting of NW-UCLA, training data includes around 5(5\%), 15(15\%), 30(30\%), 40(40\%) labeled data for each class. In Section-\ref{EE}, on the protocol of pretrain + linear evaluation of NTU RGB+D and NTU RGB+D 120, the training data includes the whole labeled data. On the protocol of pretrain + finetune of NTU RGB+D and Kinetics-Skeleton, the training data contains all labeled data.

For the encoder configuration, the used encoder of the proposed SDS-CL is DSTA~\cite{shi2020decoupled}. In Spatiotemporal-decoupling Intra-Inter Attention (SIIA), the number of multi-head $S$ is set to 4 based on experience. The semi-supervised learning process of SDS-CL is first pre-training with unlabeled data and then fine-tuning with labeled data. In the pre-training phase, the batchsize, Nesterov momentum, initial learning rate, weight decay, and warmup epoch~\cite{he2016deep} for the stochastic gradient descent (SGD) are set to 16, 0.9, 0.001, 0.0005, and 5 on both NTU RGB+D and NW-UCLA. On NTU RGB+D, the total number of training epochs is set to 70, and the learning rate is reduced by multiplying it by 0.1 after 60 epochs. On NW-UCLA, the total number of training epochs is set to 50, and the learning rate is reduced by multiplying it by 0.1 after 40 epochs. Temperature parameter $\tau$ is set to 0.07 in the formulas of contrastive loss. In the fine-tuning phase, the batchsize, initial learning rate, warmup epoch, and total epoch for the SGD optimizer with Nesterov momentum 0.9 and weight decay 0.0005 are set to 32/4, 0.1/0.02, 5/20, and 120/200 on NTU RGB+D/NW-UCLA, respectively. The experiments in the protocol of pretrain + linear evaluation/finetune contains pretrain phase and linear evaluation/finetune phase. In the pretrain phase, the batchsize, Nesterov momentum, initial learning rate, weight decay, warmup epoch, and total epochs for the SGD are set to 64, 0.9, 0.001, 0.0005, 5, and 20 on NTU RGB+D, NTU RGB+D 120, and Kinetics-Skeleton. In the linear evaluation phase, the batchsize, Nesterov momentum, initial learning rate, and total epochs for the SGD are set to 128, 0.9, 1.0, and 100 on NTU RGB+D and NTU RGB+D 120. In the finetune phase, the batchsize, Nesterov momentum, initial learning rate, weight decay, warmup epoch, and total epochs for the SGD are set to 32, 0.9, 0.1, 0.0005, 5, and 120 on NTU RGB+D and Kinetics-Skeleton. All experiments are performed via the PyTorch deep learning framework on the Linux server equipped with Titan RTX GPU.

\subsection{Experimental Result and Analysis}
We compare the proposed SDS-CL with representative methods for semi-supervised skeleton-based action recognition and show the accuracy comparison of different methods on the NTU RGB+D and NW-UCLA datasets, as shown in Table~\ref{NTU} and Table~\ref{NW-UCLA}, respectively. 

In Table~\ref{NTU}, SDS-CL is compared with some competitive semi-supervised and unsupervised methods in terms of the semi-supervised skeleton-based action recognition task under the CS and CV benchmark on the NTU RGB+D dataset. These semi-supervised methods include S$^{4}$L \cite{zhai2019s4l}, Pseudolabels \cite{lee2013pseudo}, VAT \cite{miyato2018virtual}, VAT+EntMin \cite{grandvalet2005semi}, ASSL \cite{si2020adversarial}, AL+K \cite{li2020sparse}, CD-JBF-GCN \cite{tu2022joint}. Especially, it can be seen that the proposed SDS-CL outperforms the state-of-the-art semi-supervised method (i.e., CD-JBF-GCN~\cite{tu2022joint}) on all settings. These unsupervised methods include AS-CAL \cite{rao2021augmented}, LongT GAN \cite{zheng2018unsupervised}, Holden et al. \cite{holden2015learning}, EnGAN-PoseRNN \cite{kundu2019unsupervised}, MS$^{2}$L \cite{lin2020ms2l}, Skeleton-Contrastive \cite{thoker2021skeleton}, 3s-Colorization \cite{yang2021skeleton}, 3s-CrosSCLR \cite{li20213d}, and 3s-AimCLR \cite{guo2022contrastive}. In particular, with 10\% labeled data of training sets, the proposed SDS-CL gains 2.8\% improvement compared with 3s-CrosSCLR on CS benchmark, and 4.1\% improvement compared with 3s-Colorization on CV benchmark. And SDS-CL is comparable to AimCLR, in which SDS-CL performs better on CV (with 10\% labeled data) while AimCLR performs better on CS (with 10\% labeled data).

In Table~\ref{NW-UCLA}, it can be seen that the proposed SDS-CL outperforms the previous semi-supervised methods (i.e., S$^{4}$L \cite{zhai2019s4l}, Pseudolabels \cite{lee2013pseudo}, VAT \cite{miyato2018virtual}, VAT+EntMin \cite{grandvalet2005semi}, ASSL \cite{si2020adversarial}, and AL+K \cite{li2020sparse}) and unsupervised method (i.e., MS$^{2}$L \cite{lin2020ms2l}) on all settings of NW-UCLA, indicating the effectiveness of SDS-CL. Especially, the performance of the proposed SDS-CL exceeds that of the state-of-the-art semi-supervised method (i.e., AL+K~\cite{li2020sparse}) by 3.9\% on the setting of 40\% labeled data. 

In particular, as the contrastive learning-based method, the proposed SDS-CL performs better than the SOTA contrastive learning-based methods, i.e., 3s-Colorization \cite{yang2021skeleton} on NTU RGB+D with CV benchmark, 3s-CrosSCLR \cite{li20213d} on NTU RGB+D with CS benchmark,  and MS$^2$L~\cite{lin2020ms2l} on NW-UCLA. It is illustrated that the proposed SDS-CL is an effective contrastive learning method.


\subsection{Qualitative Analysis}
\begin{figure*}
    \centering
    \includegraphics[width=7.1in]{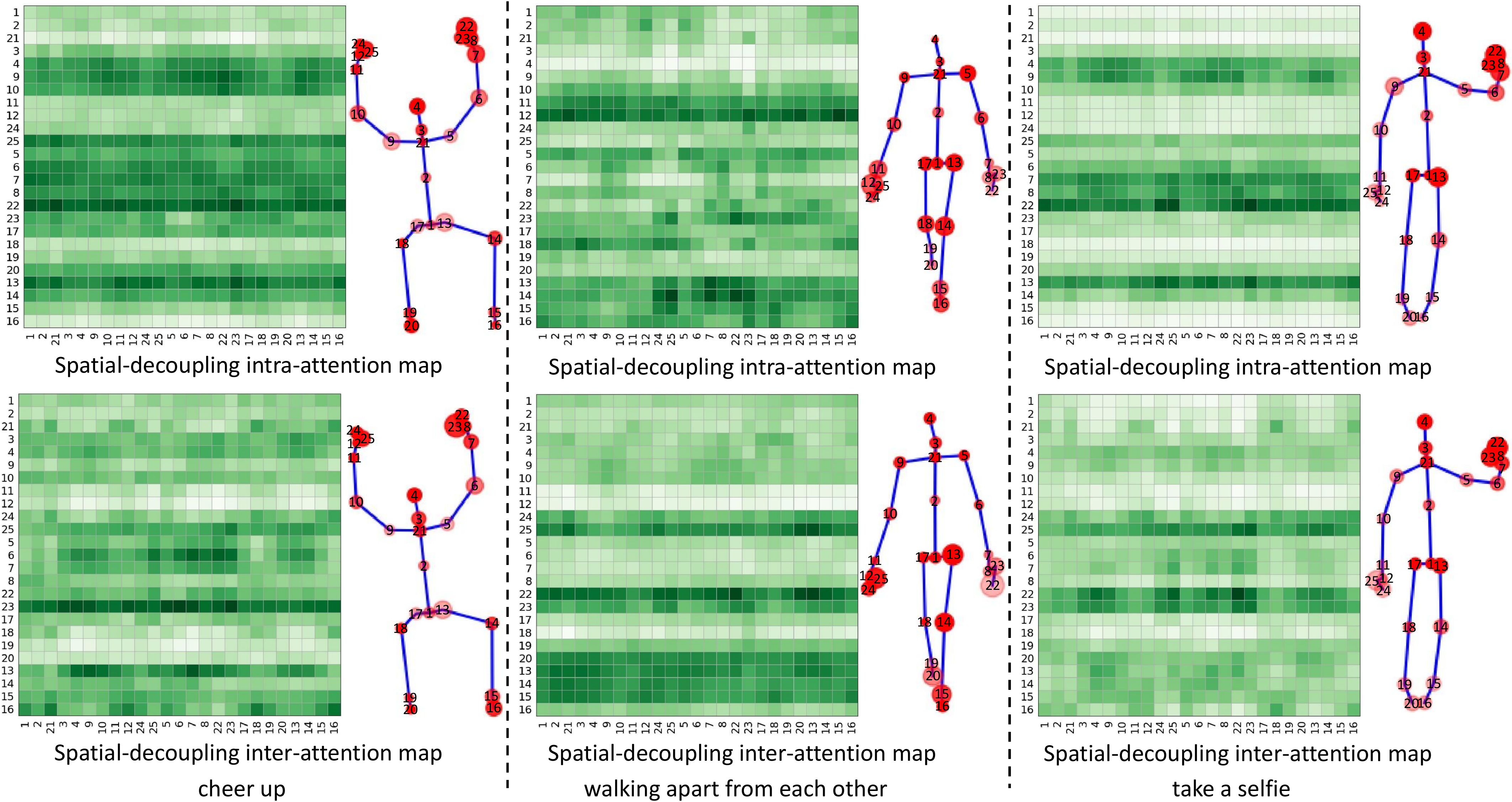}
    \caption{The visualization of the spatial-decoupling intra-attention/inter-attention map in J-SIIA for different actions (including ``cheer up", ``walking apart from each other", ``take a selfie"). In each group, the first and second rows display spatial-decoupling intra-attention and inter-attention maps, respectively. Each row includes the attention map on the left and the corresponding skeleton visualization on the right. The joints with the darker color of row-elements in attention map denotes that they are key to the action. The larger sizes of the red circle in the skeleton visualization indicate the corresponding joints play an important role in the skeleton action. For better view, please see ×3 original color PDF.}
    \label{spatial_attention}
\end{figure*}
\begin{figure*}
    \centering
    \includegraphics[width=5.8in]{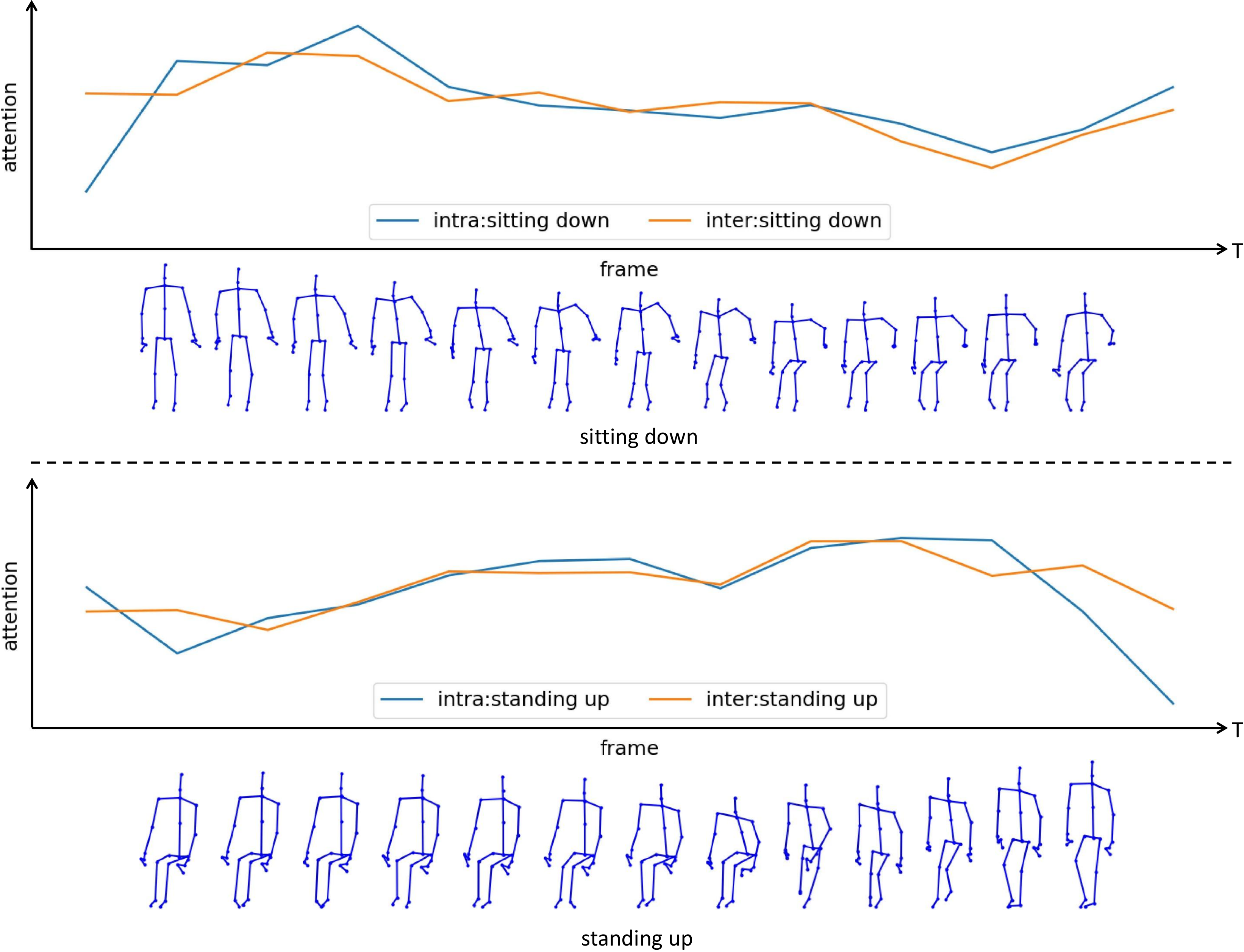}
    \caption{The visualization of the temporal-decoupling intra-attention/inter-attention map in J-SIIA for different actions (including ``sitting down", and ``standing up"). The above displays the skeleton sequences, and the bottom displays their attention values over time.}
    \label{temporal_attention}
\end{figure*}

\begin{table}[!t]
\renewcommand{\arraystretch}{1.3}
\caption{Accuracies (\%) obtained by SDS-CL with different attention mechanisms on NTU RGB+D (CS) with 5\% labeled data.}
\label{SIIA}
\centering
\begin{tabular}{c|c|c||c}
\hline\hline
\multirow{2}{*}{Baseline} & \multicolumn{2}{c||}{SIIA} & \multirow{2}{*}{Accuracy} \\ \cline{2-3}
& Spatial-decoupling & Temporal-decoupling & \\ \hline
A1 & $\times$ & $\times$ & 68.6  \\ \hline
A2 & $\surd$ & $\times$ & 70.0  \\ \hline
A3 & $\times$ & $\surd$ & 70.1  \\ \hline
A4 & $\surd$ & $\surd$ & {\bf 71.3} \\ \hline
\end{tabular}
\end{table}

J-SIIA and M-SIIA in SIIA aim to respectively learn spatial- and temporal-specific information in joint and motion modalities by calculating spatial- and temporal-decoupling intra-attention/inter-attention maps. Considering the important roles of such attentions, we visualize the learned spatial- and temporal-decoupling intra-/inter-attentions to verify their ability to capture spatial- and temporal-specific information on NTU RGB+D.
\subsubsection{Spatial-decoupling Intra-inter Attention}
Figure~\ref{spatial_attention} shows spatial-decoupling intra-attention and inter-attention for different actions. In each group, the first and second rows display spatial-decoupling intra-attention maps and inter-attention maps, respectively. Each row includes an attention map on the left and the corresponding skeleton visualization on the right. By row-wise summing the attention map, we can find which joints are key to the action. The larger sizes of the red circle in skeleton visualization indicate the corresponding joints play an important role in action. It can be found that: a) For the actions ``cheer up" and ``take a selfie", the joints of upper limb (especially hands) have the larger red circle, which illustrates that these joints deeply participate in these actions; b) For the action ``walking apart from each other" action, the joints of lower limb have the larger red circle, which illustrates that these joints deeply participate in these actions; c) The key joints captured by intra-attention and inter-attention come from the similar body part, which indicates the consistency of the learned spatial-decoupling intra-attentive and inter-attentive features.
\subsubsection{Temporal-decoupling Intra-Inter Attention}
Figure~\ref{temporal_attention} shows temporal-decoupling intra-attention and inter-attention for different actions. In each group, the bottom displays the skeleton sequence, and the top displays their intra-attention and inter-attention at each time. It can be found that: a) For the action ``sitting down", there is larger attention in the early process from standing to sitting; b) For the action ``standing up", there is larger attention in the later process from sitting to standing; c) For the same action, the variation trends of intra-attention and inter-attention are very similar, which also indicates that the learned spatial-decoupling intra-attentive and inter-attentive features are consistent.

\subsection{Ablation Studies}
As the main modules of the proposed X-CAR, SIIA, TSL, TSL and GL are also insightful in this work. To evaluate the
superiority of SIIA, TSL, TSL and GL, we conduct two groups of ablation studies on the NTU RGB+D (CS) with 5\% labeled data.
\subsubsection{Effect of SIIA}
Since SIIA (including J-SIIA and M-SIIA) contains the spatial-decoupling intra-inter attention and temporal-decoupling intra-inter attention, we set four baselines as follows,
\begin{description}
    \item[{\bf A1}] SDS-CL without SIIA. 
    \item[{\bf A2}] SDS-CL without temporal-decoupling intra-inter attention, with spatial-decoupling intra-inter attention. This can be seen as one baseline version of SIIA.
    \item[{\bf A3}] SDS-CL without spatial-decoupling intra-inter attention, with temporal-decoupling intra-inter attention. This can be seen as one baseline version of SIIA.
    \item[{\bf A4}] SDS-CL with SIIA. 
\end{description}
The recognition accuracies obtained by different baselines are shown in Table~\ref{SIIA}. Based on A1, the improvement gained by A2 and A3 is 1.4\% and 1.5\% respectively. This proves that either spatial-decoupling intra-inter attention or temporal-decoupling intra-inter attention is beneficial for the recognition task.
In A4, combining both spatial-decoupling intra-inter attention and temporal-decoupling intra-inter attention, namely SIIA, achieves the best performance (improves by 2.7\% compared with A1), which illustrates the superiority of SIIA equipped into the proposed SDS-CL in semi-supervised action recognition task.

\subsubsection{Effect of STL, TSL, and GL}
To evaluate the effectiveness of STL, TSL, and GL in SDS-CL, we set eight baselines as follows,
\begin{description}
    \item[{\bf B1}] Supervised-only training for Encoder and recognition head, and without any contrastive losses. This aims to test the basic performance by only using labeled data as training.
    \item[{\bf B2}] SDS-CL with STL, without TSL and GL, namely pre-training by only contrasting the spatial-squeezing features between joints and motions at the frame level. This aims to test the superiority of STL.
    \item[{\bf B3}] SDS-CL with TSL, without STL and GL, namely pre-training by only contrasting the temporal-squeezing features between joints and motions at the joint level. This aims to test the superiority of TSL.
    \item[{\bf B4}] SDS-CL with GL, without STL and TSL, namely pre-training by only contrasting the global features between joints and motions at the skeleton level. This aims to test the superiority of GL.
    \item[{\bf B5}] SDS-CL with STL and TSL, without GL, similar to combining B2 and B3.
    \item[{\bf B6}] SDS-CL with STL and GL, without TSL, similar to combining B2 and B4.
    \item[{\bf B7}] SDS-CL with TSL and GL, without STL, similar to combining B3 and B4.
    \item[{\bf B8}] SDS-CL with STL, TSL, and GL, similar to combining B2, B3, and B4.
\end{description}

The recognition accuracies obtained by different baselines are shown in Table~\ref{con_loss}. Here, B2, B3, and B4 with pre-training by the single contrastive loss outperform B1 with supervised-only training, which indicates STL, TSL, and GL are effective to improve the recognition performance to some extent. 
Meanwhile, B6 (with global contrast, and temporal-specific contrast), B7 (with global contrast, and spatial-specific contrast), and B8 (with global contrast, spatial-specific contrast, and temporal-specific contrast) perform better than B4 (only with global contrast). It is proved that spatial-specific information and temporal-specific information are beneficial to learn action representations. Obviously, combining two or more contrastive losses can achieve better performance. Finally, B8 (namely SDS-CL) of combining STL, TSL, and GL (with the accuracy of 71.3\%) improves by 5\% over B1 (with accuracy of 66.3\%) without contrastive loss.
\begin{table}[!t]
\renewcommand{\arraystretch}{1.3}
\caption{Accuracies (\%) obtained by SDS-CL with different contrastive loss combinations on NTU RGB+D (CS) with 5\% labeled data.}
\label{con_loss}
\centering
\begin{tabular}{c|c|c|c||c}
\hline\hline
Baseline & STL & TSL & GL & Accuracy \\ \hline
B1 & $\times$ & $\times$ & $\times$ & 66.3  \\ \hline
B2 & $\surd$ & $\times$ & $\times$ & 69.8  \\ \hline
B3 & $\times$ & $\surd$ & $\times$ & 69.4  \\ \hline
B4 & $\times$ & $\times$ & $\surd$ & 69.9  \\ \hline
B5 & $\surd$ & $\surd$ & $\times$ & 70.3  \\ \hline
B6 & $\surd$ & $\times$ & $\surd$ & 70.5  \\ \hline
B7 & $\times$ & $\surd$ & $\surd$ & 70.8  \\ \hline
B8 (Ours) & $\surd$ & $\surd$ & $\surd$ & {\bf 71.3} \\ \hline
\end{tabular}
\end{table}

\subsubsection{Effect of different encoders}
To investigate the performance of SDS-CL with different encoders when comparing previous methods, we replace the encoder with LSTM~\cite{hochreiter1997long}/ST-GCN~\cite{yan2018spatial} to test the recognition performance on NTU RGB+D with 10\% labeled data, as shown in Table~\ref{encoder}. We can find that: 1) When using LSTM as the encoder, SDS-CL performs better than AS-CAL~\cite{rao2021augmented}; 2) When using either ST-GCN or DSTA as the encoder, SDS-CL performs better than CrosSCLR~\cite{li20213d}. This demonstrates the generalization of SDS-CL adapting to various encoders.

\begin{table}[!t]
\renewcommand{\arraystretch}{1.3}
\caption{Accuracies (\%) obtained by SDS-CL with different encoder on NTU RGB+D with 10\% labeled data. }
\label{encoder}
\centering
\begin{tabular}{l|l|c|c}
\hline\hline
Encoder & Method & CS & CV \\ \hline
\multirow{2}{*}{LSTM} & AS-CAL \cite{rao2021augmented} & 52.2 & 57.3 \\ \cline{2-4}
& SDS-CL & 62.7 & 68.5 \\ \hline
\multirow{2}{*}{ST-GCN} & CrosSCLR \cite{li20213d} & 70.5 & 75.9 \\ \cline{2-4}
& SDS-CL & 72.8 & 77.2 \\ \hline
\multirow{2}{*}{DSTA} & CrosSCLR \cite{li20213d} & 72.9 & 75.8 \\ \cline{2-4}
& SDS-CL (Ours) &  \textbf{76.0} & \textbf{82.6} \\ \hline
\end{tabular}
\end{table}

\subsection{Extensive Experiment}
\label{EE}
To more fairly demonstrate the superiority of the proposed SDS-CL, we also conduct the comparative experiments in additional two protocols, i.e., pretrain + linear evaluation, as well as pretrain + finetune.
\subsubsection{Pretrain + Linear Evaluation}
In the protocol of pretrain + linear evaluation, the proposed SDS-CL is conducted the unsupervised pre-training with unlabeled data first, and then a linear classifier is performed the supervised training with labeled data by freezing pre-trained encoder. The performance comparisons between the proposed SDS-CL and the other competitive methods on NTU RGB+D and NTU RGB+D 120~\cite{liu2019ntu} (these two datasets are widely used to test this protocol) are shown in Table~\ref{linear_evaluation-NTU} and~\ref{linear_evaluation-NTU120}, respectively. It can be found that SDS-CL shows the competitive performance, which is comparable to the SOTA method, i.e., CrosSCLR~\cite{li20213d}, and is better than the rest alternatives. 

\begin{table}[!t]
\renewcommand{\arraystretch}{1.3}
\caption{Recognition accuracies (\%) obtained by different methods on NTU RGB+D in the protocol of pretrain + linear evaluation.}
\label{linear_evaluation-NTU}
\centering
\begin{tabular}{l||c|c}
\hline\hline
Method & CS & CV \\ \hline
LongT GAN \cite{zheng2018unsupervised} & 39.1 & 48.1 \\ \hline
P\&C \cite{su2020predict} & 50.7 & 76.3 \\ \hline
MS$^{2}$L \cite{lin2020ms2l} & 52.6 & - \\ \hline
PCRP \cite{xu2020prototypical} & 54.9 & 63.4 \\ \hline
AS-CAL \cite{rao2021augmented} & 58.5 & 64.8 \\ \hline
Tanfous et al.\cite{tanfous2022and} & 67.0 & 76.3 \\ \hline
CRRL \cite{wang2022contrast} & 67.6 & 73.8 \\ \hline
EnGAN-PoseRNN \cite{kundu2019unsupervised} & 68.6 & 77.8 \\ \hline
SeBiReNet \cite{nie2020unsupervised} & - & 79.7 \\ \hline
‘TS’ Colorization \cite{yang2021skeleton} & 71.6 & 79.9 \\ \hline
CrosSCLR \cite{li20213d} & 72.9 & \textbf{79.9} \\ \hline
SDS-CL (Ours) & \textbf{73.6} & 78.9 \\ \hline
\end{tabular}
\end{table}

\begin{table}[!t]
\renewcommand{\arraystretch}{1.3}
\caption{Recognition accuracies (\%) obtained by different methods on NTU RGB+D 120 in the protocol of pretrain + linear evaluation.}
\label{linear_evaluation-NTU120}
\centering
\begin{tabular}{l||c|c}
\hline\hline
Method & CS & CE \\ \hline
LongT GAN \cite{zheng2018unsupervised} & 35.6 & 39.7 \\ \hline
P\&C \cite{su2020predict} & 42.7 & 41.7 \\ \hline
PCRP \cite{xu2020prototypical} & 43.0 & 44.6 \\ \hline
AS-CAL \cite{rao2021augmented} & 48.6 & 49.2 \\ \hline 
CrosSCLR \cite{li20213d} & \textbf{50.9} & 46.7 \\ \hline
SDS-CL (Ours) & 50.6 & \textbf{55.6} \\ \hline
\end{tabular}
\end{table}

\subsubsection{Pretrain + Finetune}
In the protocol of pretrain + finetune, the proposed SDS-CL is first pre-trained by three-combined contrastive losses (i.e., STL, GL, and TSL) on all training data without labels, and then SDS-CL with a linear classifier is fine-tuned in a fully supervised way. The performance comparisons between SDS-CL and the other competitive methods on NTU RGB+D and Kinetics-Skeleton (these two datasets are widely used to test this protocol) are shown in Table~\ref{finetuned-ntu}, and~\ref{finetuned-kinetics}, respectively. In Table~\ref{finetuned-ntu}, for a fair comparison, similar to 3s-CrosSCLR~\cite{li20213d}, 3s-AimCLR~\cite{guo2022contrastive}, and 3s-Colorization~\cite{yang2021skeleton}, SDS-CL also learns effective information from three modalities (i.e., joint, motion, and bone), named as 3s-SDS-CL. 3s-SDS-CL is comparable to the SOTA method, i.e., IIP-Transformer~\cite{wang2021iip}, and performs better than the rest comparative methods. In particular, 3s-SDS-CL with the accuracy of 91.9\%/97.2\% improves by 3.9\%/2.3\% over 3s-Colorization~\cite{yang2021skeleton} with the accuracy of 88.0\%/94.9\% on CS/CV, where both of them consider the joint, motion, and bone modals. In Table~\ref{finetuned-kinetics}, we validate the effectiveness of SDS-CL on a more accessible yet more challenging 2D skeleton dataset, e.g., Kinetics-Skeleton dataset. 
As shown in Table~\ref{finetuned-kinetics}, the performance between SDS-CL and GCN-NAS (SOTA method) is comparable. This proves that SDS-CL can also perform well on 2D skeleton sequences.

\begin{table}[!t]
\renewcommand{\arraystretch}{1.3}
\caption{Recognition accuracies (\%) obtained by different methods on NTU RGB+D in the protocol of pretrain + finetune.}
\label{finetuned-ntu}
\centering
\begin{tabular}{l||c|c}
\hline\hline
Method & CS & CV \\ \hline
Li et al. \cite{li2018unsupervised} & 63.9 & 68.1 \\ \hline
MS$^{2}$L \cite{lin2020ms2l} & 78.6 & - \\ \hline
ST-GCN \cite{yan2018spatial} & 81.5 & 88.3 \\ \hline
3s-CrosSCLR \cite{li20213d} & 86.2 & 92.5 \\ \hline
3s-AimCLR \cite{guo2022contrastive} & 86.9 & 92.8 \\ \hline
3s-Colorization \cite{yang2021skeleton} & 88.0 & 94.9 \\ \hline
2s-AGCN \cite{shi2019two} & 88.5 & 95.1 \\ \hline
AGC-LSTM \cite{si2019attention} & 89.2 & 95.0 \\ \hline
Shift-GCN \cite{cheng2020skeleton} & 90.7 & 96.5 \\ \hline
MS-G3D \cite{liu2020disentangling} & 91.5 & 96.2 \\ \hline
DSTA \cite{shi2020decoupled} & 91.5 & 96.4 \\ \hline
IIP-Transformer \cite{wang2021iip} & \textbf{92.3} & 96.4 \\ \hline
3s-SDS-CL (Ours) & 91.9 & \textbf{97.2} \\ \hline
\end{tabular}
\end{table}

\begin{table}[!t]
\renewcommand{\arraystretch}{1.3}
\caption{Recognition accuracies (\%) obtained by different methods on Kinetics-Skeleton in the protocol of pretrain + finetune.}
\label{finetuned-kinetics}
\centering
\begin{tabular}{l||c}
  \hline
  Method & Accuracy \\ \hline\hline
  Deep LSTM~\cite{shahroudy2016ntu} & 16.4 \\ \hline
  TCN~\cite{kim2017interpretable} & 20.3 \\ \hline
  ST-GCN~\cite{yan2018spatial} & 30.7 \\ \hline
  AS-GCN~\cite{li2019actional} & 34.8 \\ \hline
  2s-AGCN~\cite{shi2019two} & 36.1 \\ \hline
  DGNN~\cite{shi2019skeleton} & 36.9 \\ \hline
  GCN-NAS~\cite{peng2020learning} & 37.1 \\ \hline
  SDS-CL (Ours) & \textbf{37.3} \\ \hline
\end{tabular}
\end{table}

\section{Conclusion}
\label{conclusion}
In this work, we proposed a novel Spatiotemporal Decouple-and-Squeeze Contrastive Learning (SDS-CL) framework that jointly contrasts spatial-squeezing features, temporal-squeezing features, and global features for well addressing the problem of semi-supervised skeleton-based action recognition. Compared with the method of only contrasting global features confusing spatiotemporal information, contrasting features additionally decoupling spatial- and temporal-specific information can obtain more abundant representations. Here, the framework of the proposed SDS-CL has two main insights. First, we designed a new Spatiotemporal-decoupling Intra-Inter Attention (SIIA) that aims to separately capture spatial- and temporal-specific information by calculating spatial- and temporal-decoupling intra-attention/inter-attention maps. Second, we presented a new Spatial-squeezing Temporal-contrasting Loss (STL), a new Temporal-squeezing Spatial-contrasting Loss (TSL), and a Global-contrasting Loss (GL) that aim to promote consistency of representations by contrasting spatial-squeezing features at the frame level, temporal-squeezing features at the joint level, and global features at the skeleton level. Experimental results on four public datasets show that the proposed SDS-CL achieved the performance gains. This work relies on the location accuracy of the extracted skeletal joints. In the future, we will extend the proposed SDS-CL to be robust to location of skeletal joints even with some noise by weakly/strongly augmenting data.

\ifCLASSOPTIONcaptionsoff
  \newpage
\fi

\bibliographystyle{IEEEtran}
\bibliography{IEEEabrv,main}

\begin{thebibliography}{10}
\providecommand{\url}[1]{#1}
\csname url@samestyle\endcsname
\providecommand{\newblock}{\relax}
\providecommand{\bibinfo}[2]{#2}
\providecommand{\BIBentrySTDinterwordspacing}{\spaceskip=0pt\relax}
\providecommand{\BIBentryALTinterwordstretchfactor}{4}
\providecommand{\BIBentryALTinterwordspacing}{\spaceskip=\fontdimen2\font plus
\BIBentryALTinterwordstretchfactor\fontdimen3\font minus
  \fontdimen4\font\relax}
\providecommand{\BIBforeignlanguage}[2]{{%
\expandafter\ifx\csname l@#1\endcsname\relax
\typeout{** WARNING: IEEEtran.bst: No hyphenation pattern has been}%
\typeout{** loaded for the language `#1'. Using the pattern for}%
\typeout{** the default language instead.}%
\else
\language=\csname l@#1\endcsname
\fi
#2}}
\providecommand{\BIBdecl}{\relax}
\BIBdecl

\bibitem{weinland2011survey}
D.~Weinland, R.~Ronfard, and E.~Boyer, ``A survey of vision-based methods for
  action representation, segmentation and recognition,'' \emph{Computer Vision
  and Image Understanding}, vol. 115, no.~2, pp. 224--241, 2011.

\bibitem{poppe2010survey}
R.~Poppe, ``A survey on vision-based human action recognition,'' \emph{Image
  and Vision Computing}, vol.~28, no.~6, pp. 976--990, 2010.

\bibitem{shu2019hierarchical}
X.~Shu, J.~Tang, G.~Qi, W.~Liu, and J.~Yang, ``Hierarchical long short-term
  concurrent memory for human interaction recognition,'' \emph{IEEE
  Transactions on Pattern Analysis and Machine Intelligence}, vol.~40, no.~3,
  pp. 1110--1118, 2021.

\bibitem{shu2020host}
X.~Shu, L.~Zhang, Y.~Sun, and J.~Tang, ``Host--parasite: Graph lstm-in-lstm for
  group activity recognition,'' \emph{IEEE Transactions on Neural Networks and
  Learning Systems}, vol.~32, no.~2, pp. 663--674, 2021.

\bibitem{shu2022expansion}
X.~Shu, J.~Yang, R.~Yan, and Y.~Song, ``Expansion-squeeze-excitation fusion
  network for elderly activity recognition,'' \emph{IEEE Transactions on
  Circuits and Systems for Video Technology}, vol.~32, no.~8, pp. 5281--5292,
  2023.

\bibitem{tang2019coherence}
J.~Tang, X.~Shu, R.~Yan, and L.~Zhang, ``Coherence constrained graph lstm for
  group activity recognition,'' \emph{IEEE Transactions on Pattern Analysis and
  Machine Intelligence}, vol.~44, no.~2, pp. 636--647, 2019.

\bibitem{liu2019hidden}
X.~Liu, H.~Shi, X.~Hong, H.~Chen, D.~Tao, and G.~Zhao, ``Hidden states
  exploration for 3d skeleton-based gesture recognition,'' in \emph{IEEE Winter
  Conference on Applications of Computer Vision (WACV)}, 2019, pp. 1846--1855.

\bibitem{yu2021searching}
Z.~Yu, B.~Zhou, J.~Wan, P.~Wang, H.~Chen, X.~Liu, S.~Z. Li, and G.~Zhao,
  ``Searching multi-rate and multi-modal temporal enhanced networks for gesture
  recognition,'' \emph{IEEE Transactions on Image Processing}, vol.~30, no.~6,
  pp. 5626--5640, 2021.

\bibitem{kim2017interpretable}
T.~S. Kim and A.~Reiter, ``Interpretable 3d human action analysis with temporal
  convolutional networks,'' in \emph{IEEE Conference on Computer Vision and
  Pattern Recognition (CVPR) Workshops}, 2017, pp. 1623--1631.

\bibitem{li2018independently}
S.~Li, W.~Li, C.~Cook, C.~Zhu, and Y.~Gao, ``Independently recurrent neural
  network (indrnn): Building a longer and deeper rnn,'' in \emph{IEEE
  Conference on Computer Vision and Pattern Recognition (CVPR)}, 2018, pp.
  5457--5466.

\bibitem{shu2022spatiotemporal}
X.~Shu, L.~Zhang, G.-J. Qi, W.~Liu, and J.~Tang, ``Spatiotemporal co-attention
  recurrent neural networks for human-skeleton motion prediction,'' \emph{IEEE
  Transactions on Pattern Analysis and Machine Intelligence}, vol.~44, no.~6,
  pp. 3300--3315, 2022.

\bibitem{du2015hierarchical}
Y.~Du, W.~Wang, and L.~Wang, ``Hierarchical recurrent neural network for
  skeleton based action recognition,'' in \emph{IEEE Conference on Computer
  Vision and Pattern Recognition (CVPR)}, 2015, pp. 1110--1118.

\bibitem{zhang2017view}
P.~Zhang, C.~Lan, J.~Xing, W.~Zeng, J.~Xue, and N.~Zheng, ``View adaptive
  recurrent neural networks for high performance human action recognition from
  skeleton data,'' in \emph{IEEE International Conference on Computer Vision
  (ICCV)}, 2017, pp. 2117--2126.

\bibitem{li2018co}
C.~Li, Q.~Zhong, D.~Xie, and S.~Pu, ``Co-occurrence feature learning from
  skeleton data for action recognition and detection with hierarchical
  aggregation,'' in \emph{International Joint Conference on Artificial
  Intelligence (IJCAI)}, 2018, pp. 786--792.

\bibitem{ke2017new}
Q.~Ke, M.~Bennamoun, S.~An, F.~Sohel, and F.~Boussaid, ``A new representation
  of skeleton sequences for 3d action recognition,'' in \emph{IEEE Conference
  on Computer Vision and Pattern Recognition (CVPR)}, 2017, pp. 3288--3297.

\bibitem{yan2018spatial}
S.~Yan, Y.~Xiong, and D.~Lin, ``Spatial temporal graph convolutional networks
  for skeleton-based action recognition,'' in \emph{AAAI Conference on
  Artificial Intelligence (AAAI)}, 2018, pp. 7444--7452.

\bibitem{tang2018deep}
Y.~Tang, Y.~Tian, J.~Lu, P.~Li, and J.~Zhou, ``Deep progressive reinforcement
  learning for skeleton-based action recognition,'' in \emph{IEEE Conference on
  Computer Vision and Pattern Recognition (CVPR)}, 2018, pp. 5323--5332.

\bibitem{liu20193d}
X.~Liu and G.~Zhao, ``3d skeletal gesture recognition via sparse coding of
  time-warping invariant riemannian trajectories,'' in \emph{International
  Conference on Multimedia Modeling (MMM)}, 2019, pp. 678--690.

\bibitem{liu20203d}
X.~Liu, H.~Shi, X.~Hong, H.~Chen, D.~Tao, and G.~Zhao, ``3d skeletal gesture
  recognition via hidden states exploration,'' \emph{IEEE Transactions on Image
  Processing}, vol.~29, no.~2, pp. 4583--4597, 2020.

\bibitem{yan2017skeleton}
Y.~Yan, J.~Xu, B.~Ni, W.~Zhang, and X.~Yang, ``Skeleton-aided articulated
  motion generation,'' in \emph{ACM International Conference on Multimedia (ACM
  MM)}, 2017, pp. 199--207.

\bibitem{liu2017skeleton}
J.~Liu, A.~Shahroudy, D.~Xu, A.~C. Kot, and G.~Wang, ``Skeleton-based action
  recognition using spatio-temporal lstm network with trust gates,'' \emph{IEEE
  Transactions on Pattern Analysis and Machine Intelligence}, vol.~40, no.~12,
  pp. 3007--3021, 2017.

\bibitem{li2017skeleton}
C.~Li, Q.~Zhong, D.~Xie, and S.~Pu, ``Skeleton-based action recognition with
  convolutional neural networks,'' in \emph{IEEE International Conference on
  Multimedia \& Expo (ICEM) Workshops}, 2017, pp. 597--600.

\bibitem{li2019actional}
M.~Li, S.~Chen, X.~Chen, Y.~Zhang, Y.~Wang, and Q.~Tian, ``Actional-structural
  graph convolutional networks for skeleton-based action recognition,'' in
  \emph{IEEE/CVF Conference on Computer Vision and Pattern Recognition (CVPR)},
  2019, pp. 3595--3603.

\bibitem{shi2019two}
L.~Shi, Y.~Zhang, J.~Cheng, and H.~Lu, ``Two-stream adaptive graph
  convolutional networks for skeleton-based action recognition,'' in
  \emph{IEEE/CVF Conference on Computer Vision and Pattern Recognition (CVPR)},
  2019, pp. 12\,026--12\,035.

\bibitem{shi2020decoupled}
{L. Shi, Y. Zhang, J. Cheng, and H. Lu}, ``Decoupled spatial-temporal attention
  network for skeleton-based action-gesture recognition,'' in \emph{Asian
  Conference on Computer Vision (ACCV)}, 2020, pp. 38--53.

\bibitem{wang2021iip}
Q.~Wang, J.~Peng, S.~Shi, T.~Liu, J.~He, and R.~Weng, ``Iip-transformer:
  Intra-inter-part transformer for skeleton-based action recognition,''
  \emph{arXiv preprint arXiv:2110.13385}, 2021.

\bibitem{li20213d}
L.~Li, M.~Wang, B.~Ni, H.~Wang, J.~Yang, and W.~Zhang, ``3d human action
  representation learning via cross-view consistency pursuit,'' in
  \emph{IEEE/CVF Conference on Computer Vision and Pattern Recognition (CVPR)},
  2021, pp. 4741--4750.

\bibitem{thoker2021skeleton}
F.~M. Thoker, H.~Doughty, and C.~G. Snoek, ``Skeleton-contrastive 3d action
  representation learning,'' in \emph{ACM International Conference on
  Multimedia (ACM MM)}, 2021, pp. 1655--1663.

\bibitem{hussein2013human}
M.~E. Hussein, M.~Torki, M.~A. Gowayyed, and M.~El-Saban, ``Human action
  recognition using a temporal hierarchy of covariance descriptors on 3d joint
  locations,'' in \emph{International Joint Conference on Artificial
  Intelligence (IJCAI)}, 2013, pp. 2466--2472.

\bibitem{vemulapalli2014human}
R.~Vemulapalli, F.~Arrate, and R.~Chellappa, ``Human action recognition by
  representing 3d skeletons as points in a lie group,'' in \emph{IEEE
  Conference on Computer Vision and Pattern Recognition (CVPR)}, 2014, pp.
  588--595.

\bibitem{liu2020adversarial}
J.~Liu, N.~Akhtar, and A.~Mian, ``Adversarial attack on skeleton-based human
  action recognition,'' \emph{IEEE Transactions on Neural Networks and Learning
  Systems}, vol.~33, no.~4, pp. 1609--1622, 2022.

\bibitem{li2021memory}
C.~Li, C.~Xie, B.~Zhang, J.~Han, X.~Zhen, and J.~Chen, ``Memory attention
  networks for skeleton-based action recognition,'' \emph{IEEE Transactions on
  Neural Networks and Learning Systems}, vol.~33, no.~9, pp. 4800--4814, 2022.

\bibitem{liu2016spatio}
J.~Liu, A.~Shahroudy, D.~Xu, and G.~Wang, ``Spatio-temporal lstm with trust
  gates for 3d human action recognition,'' in \emph{European Conference on
  Computer Vision (ECCV)}, 2016, pp. 816--833.

\bibitem{shahroudy2016ntu}
A.~Shahroudy, J.~Liu, T.-T. Ng, and G.~Wang, ``Ntu rgb+ d: A large scale
  dataset for 3d human activity analysis,'' in \emph{IEEE Conference on
  Computer Vision and Pattern Recognition (CVPR)}, 2016, pp. 1010--1019.

\bibitem{du2015skeleton}
Y.~Du, Y.~Fu, and L.~Wang, ``Skeleton based action recognition with
  convolutional neural network,'' in \emph{Asian Conference on Pattern
  Recognition (ACPR)}, 2015, pp. 579--583.

\bibitem{liu2017enhanced}
M.~Liu, H.~Liu, and C.~Chen, ``Enhanced skeleton visualization for view
  invariant human action recognition,'' \emph{Pattern Recognition}, vol.~68,
  no.~8, pp. 346--362, 2017.

\bibitem{song2020stronger}
Y.-F. Song, Z.~Zhang, C.~Shan, and L.~Wang, ``Stronger, faster and more
  explainable: A graph convolutional baseline for skeleton-based action
  recognition,'' in \emph{ACM International Conference on Multimedia (ACM MM)},
  2020, pp. 1625--1633.

\bibitem{chen2021channel}
Y.~Chen, Z.~Zhang, C.~Yuan, B.~Li, Y.~Deng, and W.~Hu, ``Channel-wise topology
  refinement graph convolution for skeleton-based action recognition,'' in
  \emph{IEEE/CVF International Conference on Computer Vision (ICCV)}, 2021, pp.
  13\,359--13\,368.

\bibitem{shi2022multiscale}
H.~Shi, W.~Peng, H.~Chen, X.~Liu, and G.~Zhao, ``Multiscale 3d-shift graph
  convolution network for emotion recognition from human actions,'' \emph{IEEE
  Intelligent Systems}, vol.~37, no.~4, pp. 103--110, 2022.

\bibitem{chapelle2009semi}
O.~Chapelle, B.~Scholkopf, and A.~Zien, ``Semi-supervised learning (chapelle,
  o. et al., eds.; 2006)[book reviews],'' \emph{IEEE Transactions on Neural
  Networks}, vol.~20, no.~3, pp. 542--542, 2009.

\bibitem{liu2020semi}
H.~Liu, C.~Liu, and R.~Ding, ``Semi-supervised long short-term memory for human
  action recognition,'' \emph{The Journal of Engineering}, vol. 2020, no.~13,
  pp. 373--378, 2020.

\bibitem{si2020adversarial}
C.~Si, X.~Nie, W.~Wang, L.~Wang, T.~Tan, and J.~Feng, ``Adversarial
  self-supervised learning for semi-supervised 3d action recognition,'' in
  \emph{European Conference on Computer Vision (ECCV)}, 2020, pp. 35--51.

\bibitem{li2020sparse}
J.~Li and E.~Shlizerman, ``Sparse semi-supervised action recognition with
  active learning,'' \emph{arXiv preprint arXiv:2012.01740}, 2020.

\bibitem{li2020iterate}
{J. Li and E. Shlizerman}, ``Iterate \& cluster: Iterative semi-supervised
  action recognition,'' \emph{arXiv preprint arXiv:2006.06911}, 2020.

\bibitem{tu2022joint}
Z.~Tu, J.~Zhang, H.~Li, Y.~Chen, and J.~Yuan, ``Joint-bone fusion graph
  convolutional network for semi-supervised skeleton action recognition,''
  \emph{arXiv preprint arXiv:2202.04075}, 2022.

\bibitem{holden2015learning}
D.~Holden, J.~Saito, T.~Komura, and T.~Joyce, ``Learning motion manifolds with
  convolutional autoencoders,'' in \emph{SIGGRAPH Asia}, 2015, pp. 1--4.

\bibitem{zheng2018unsupervised}
N.~Zheng, J.~Wen, R.~Liu, L.~Long, J.~Dai, and Z.~Gong, ``Unsupervised
  representation learning with long-term dynamics for skeleton based action
  recognition,'' in \emph{AAAI Conference on Artificial Intelligence (AAAI)},
  2018, pp. 1--6.

\bibitem{kundu2019unsupervised}
J.~N. Kundu, M.~Gor, P.~K. Uppala, and V.~B. Radhakrishnan, ``Unsupervised
  feature learning of human actions as trajectories in pose embedding
  manifold,'' in \emph{Winter Conference on Applications of Computer Vision
  (WACV)}, 2019, pp. 1459--1467.

\bibitem{lin2020ms2l}
L.~Lin, S.~Song, W.~Yang, and J.~Liu, ``Ms2l: Multi-task self-supervised
  learning for skeleton based action recognition,'' in \emph{ACM International
  Conference on Multimedia (ACM MM)}, 2020, pp. 2490--2498.

\bibitem{su2020predict}
K.~Su, X.~Liu, and E.~Shlizerman, ``Predict \& cluster: Unsupervised skeleton
  based action recognition,'' in \emph{IEEE/CVF Conference on Computer Vision
  and Pattern Recognition (CVPR)}, 2020, pp. 9631--9640.

\bibitem{xu2020prototypical}
S.~Xu, H.~Rao, X.~Hu, and B.~Hu, ``Prototypical contrast and reverse
  prediction: Unsupervised skeleton based action recognition,'' \emph{arXiv
  preprint arXiv:2011.07236}, 2020.

\bibitem{rao2021augmented}
H.~Rao, S.~Xu, X.~Hu, J.~Cheng, and B.~Hu, ``Augmented skeleton based
  contrastive action learning with momentum lstm for unsupervised action
  recognition,'' \emph{Information Sciences}, vol. 569, pp. 90--109, 2021.

\bibitem{gao2021contrastive}
X.~Gao, Y.~Yang, and S.~Du, ``Contrastive self-supervised learning for skeleton
  action recognition,'' in \emph{Neural Information Processing Systems
  Workshops (NeurIPS)}, 2021, pp. 51--61.

\bibitem{su2021self}
Y.~Su, G.~Lin, and Q.~Wu, ``Self-supervised 3d skeleton action representation
  learning with motion consistency and continuity,'' in \emph{IEEE/CVF
  International Conference on Computer Vision (ICCV)}, 2021, pp.
  13\,328--13\,338.

\bibitem{yang2021skeleton}
S.~Yang, J.~Liu, S.~Lu, M.~H. Er, and A.~C. Kot, ``Skeleton cloud colorization
  for unsupervised 3d action representation learning,'' in \emph{IEEE/CVF
  International Conference on Computer Vision (ICCV)}, 2021, pp.
  13\,423--13\,433.

\bibitem{tanfous2022and}
A.~B. Tanfous, A.~Zerroug, D.~Linsley, and T.~Serre, ``How and what to learn:
  Taxonomizing self-supervised learning for 3d action recognition.'' in
  \emph{IEEE/CVF Winter Conference on Applications of Computer Vision (WACV)},
  2022, pp. 2888--2897.

\bibitem{liu2021imigue}
X.~Liu, H.~Shi, H.~Chen, Z.~Yu, X.~Li, and G.~Zhao, ``imigue: An identity-free
  video dataset for micro-gesture understanding and emotion analysis,'' in
  \emph{IEEE/CVF Conference on Computer Vision and Pattern Recognition (CVPR)},
  2021, pp. 10\,631--10\,642.

\bibitem{guo2022contrastive}
T.~Guo, H.~Liu, Z.~Chen, M.~Liu, T.~Wang, and R.~Ding, ``Contrastive learning
  from extremely augmented skeleton sequences for self-supervised action
  recognition,'' in \emph{AAAI Conference on Artificial Intelligence (AAAI)},
  2022, pp. 762--770.

\bibitem{wu2018unsupervised}
Z.~Wu, Y.~Xiong, S.~X. Yu, and D.~Lin, ``Unsupervised feature learning via
  non-parametric instance discrimination,'' in \emph{IEEE Conference on
  Computer Vision and Pattern Recognition (CVPR)}, 2018, pp. 3733--3742.

\bibitem{misra2020self}
I.~Misra and L.~v.~d. Maaten, ``Self-supervised learning of pretext-invariant
  representations,'' in \emph{IEEE/CVF Conference on Computer Vision and
  Pattern Recognition (CVPR)}, 2020, pp. 6707--6717.

\bibitem{henaff2020data}
O.~Henaff, ``Data-efficient image recognition with contrastive predictive
  coding,'' in \emph{International Conference on Machine Learning (ICML)},
  2020, pp. 4182--4192.

\bibitem{chen2020improved}
X.~Chen, H.~Fan, R.~Girshick, and K.~He, ``Improved baselines with momentum
  contrastive learning,'' \emph{arXiv preprint arXiv:2003.04297}, 2020.

\bibitem{he2020momentum}
K.~He, H.~Fan, Y.~Wu, S.~Xie, and R.~Girshick, ``Momentum contrast for
  unsupervised visual representation learning,'' in \emph{IEEE/CVF Conference
  on Computer Vision and Pattern Recognition (CVPR)}, 2020, pp. 9729--9738.

\bibitem{tian2020contrastive}
Y.~Tian, D.~Krishnan, and P.~Isola, ``Contrastive multiview coding,'' in
  \emph{European Conference on Computer Vision (ECCV)}, 2020, pp. 776--794.

\bibitem{chen2020simple}
T.~Chen, S.~Kornblith, M.~Norouzi, and G.~Hinton, ``A simple framework for
  contrastive learning of visual representations,'' in \emph{International
  Conference on Machine Learning (ICML)}, 2020, pp. 1597--1607.

\bibitem{caron2020unsupervised}
M.~Caron, I.~Misra, J.~Mairal, P.~Goyal, P.~Bojanowski, and A.~Joulin,
  ``Unsupervised learning of visual features by contrasting cluster
  assignments,'' in \emph{Neural Information Processing Systems Workshops
  (NeurIPS)}, 2020, pp. 9912--9924.

\bibitem{grill2020bootstrap}
J.-B. Grill, F.~Strub, F.~Altch{\'e}, C.~Tallec, P.~Richemond, E.~Buchatskaya,
  C.~Doersch, B.~Avila~Pires, Z.~Guo, M.~Gheshlaghi~Azar \emph{et~al.},
  ``Bootstrap your own latent-a new approach to self-supervised learning,'' in
  \emph{Neural Information Processing Systems Workshops (NeurIPS)}, 2020, pp.
  21\,271--21\,284.

\bibitem{vaswani2017attention}
A.~Vaswani, N.~Shazeer, N.~Parmar, J.~Uszkoreit, L.~Jones, A.~N. Gomez,
  {\L}.~Kaiser, and I.~Polosukhin, ``Attention is all you need,'' in
  \emph{Advances in neural information processing systems (NeurIPS)}, 2017, pp.
  5998--6008.

\bibitem{khan2021transformers}
S.~Khan, M.~Naseer, M.~Hayat, S.~W. Zamir, F.~S. Khan, and M.~Shah,
  ``Transformers in vision: A survey,'' \emph{ACM computing surveys (CSUR)},
  vol.~54, no. 10s, pp. 1--41, 2022.

\bibitem{han2022survey}
K.~Han, Y.~Wang, H.~Chen, X.~Chen, J.~Guo, Z.~Liu, Y.~Tang, A.~Xiao, C.~Xu,
  Y.~Xu \emph{et~al.}, ``A survey on vision transformer,'' \emph{IEEE
  Transactions on Pattern Analysis and Machine Intelligence}, vol.~45, no.~1,
  pp. 87--110, 2023.

\bibitem{lin2021survey}
T.~Lin, Y.~Wang, X.~Liu, and X.~Qiu, ``A survey of transformers,'' \emph{arXiv
  preprint arXiv:2106.04554}, 2021.

\bibitem{selva2022video}
J.~Selva, A.~S. Johansen, S.~Escalera, K.~Nasrollahi, T.~B. Moeslund, and
  A.~Clap{\'e}s, ``Video transformers: A survey,'' \emph{arXiv preprint
  arXiv:2201.05991}, 2022.

\bibitem{qiu2022spatio}
H.~Qiu, B.~Hou, B.~Ren, and X.~Zhang, ``Spatio-temporal tuples transformer for
  skeleton-based action recognition,'' \emph{arXiv preprint arXiv:2201.02849},
  2022.

\bibitem{shi2021star}
F.~Shi, C.~Lee, L.~Qiu, Y.~Zhao, T.~Shen, S.~Muralidhar, T.~Han, S.-C. Zhu, and
  V.~Narayanan, ``Star: Sparse transformer-based action recognition,''
  \emph{arXiv preprint arXiv:2107.07089}, 2021.

\bibitem{wang2014cross}
J.~Wang, X.~Nie, Y.~Xia, Y.~Wu, and S.-C. Zhu, ``Cross-view action modeling,
  learning and recognition,'' in \emph{IEEE Conference on Computer Vision and
  Pattern Recognition (CVPR)}, 2014, pp. 2649--2656.

\bibitem{liu2019ntu}
J.~Liu, A.~Shahroudy, M.~Perez, G.~Wang, L.-Y. Duan, and A.~C. Kot, ``Ntu rgb+
  d 120: A large-scale benchmark for 3d human activity understanding,''
  \emph{IEEE Transactions on Pattern Analysis and Machine Intelligence},
  vol.~42, no.~10, pp. 2684--2701, 2019.

\bibitem{kay2017kinetics}
W.~Kay, J.~Carreira, K.~Simonyan, B.~Zhang, C.~Hillier, S.~Vijayanarasimhan,
  F.~Viola, T.~Green, T.~Back, P.~Natsev \emph{et~al.}, ``The kinetics human
  action video dataset,'' \emph{arXiv preprint arXiv:1705.06950}, 2017.

\bibitem{cao2017realtime}
Z.~Cao, T.~Simon, S.-E. Wei, and Y.~Sheikh, ``Realtime multi-person 2d pose
  estimation using part affinity fields,'' in \emph{IEEE Conference on Computer
  Vision and Pattern Recognition (CVPR)}, 2017, pp. 7291--7299.

\bibitem{zhai2019s4l}
X.~Zhai, A.~Oliver, A.~Kolesnikov, and L.~Beyer, ``S4l: Self-supervised
  semi-supervised learning,'' in \emph{IEEE/CVF International Conference on
  Computer Vision (ICCV)}, 2019, pp. 1476--1485.

\bibitem{lee2013pseudo}
D.-H. Lee \emph{et~al.}, ``Pseudo-label: The simple and efficient
  semi-supervised learning method for deep neural networks,'' in
  \emph{International Council for Machinery Lubrication (ICML) Workshops},
  2013, p. 896.

\bibitem{miyato2018virtual}
T.~Miyato, S.-i. Maeda, M.~Koyama, and S.~Ishii, ``Virtual adversarial
  training: a regularization method for supervised and semi-supervised
  learning,'' \emph{IEEE Transactions on Pattern Analysis and Machine
  Intelligence}, vol.~41, no.~8, pp. 1979--1993, 2018.

\bibitem{grandvalet2005semi}
Y.~Grandvalet and Y.~Bengio, ``Semi-supervised learning by entropy
  minimization,'' in \emph{Advances in Neural Information Processing Systems
  (NeurIPS)}, 2004, pp. 281--296.

\bibitem{he2016deep}
K.~He, X.~Zhang, S.~Ren, and J.~Sun, ``Deep residual learning for image
  recognition,'' in \emph{IEEE Conference on Computer Vision and Pattern
  Recognition (CVPR)}, 2016, pp. 770--778.

\bibitem{hochreiter1997long}
S.~Hochreiter and J.~Schmidhuber, ``Long short-term memory,'' \emph{Neural
  computation}, vol.~9, no.~8, pp. 1735--1780, 1997.

\bibitem{wang2022contrast}
P.~Wang, J.~Wen, C.~Si, Y.~Qian, and L.~Wang, ``Contrast-reconstruction
  representation learning for self-supervised skeleton-based action
  recognition,'' \emph{IEEE Transactions on Image Processing}, vol.~31, no.~2,
  pp. 6224--6238, 2022.

\bibitem{nie2020unsupervised}
Q.~Nie, Z.~Liu, and Y.~Liu, ``Unsupervised 3d human pose representation with
  viewpoint and pose disentanglement,'' in \emph{European Conference on
  Computer Vision (ECCV)}, 2020, pp. 102--118.

\bibitem{li2018unsupervised}
J.~Li, Y.~Wong, Q.~Zhao, and M.~S. Kankanhalli, ``Unsupervised learning of
  view-invariant action representations,'' in \emph{Advances in neural
  information processing systems (NeurIPS)}, 2018, pp. 1254--1264.

\bibitem{si2019attention}
C.~Si, W.~Chen, W.~Wang, L.~Wang, and T.~Tan, ``An attention enhanced graph
  convolutional lstm network for skeleton-based action recognition,'' in
  \emph{IEEE/CVF Conference on Computer Vision and Pattern Recognition (CVPR)},
  2019, pp. 1227--1236.

\bibitem{cheng2020skeleton}
K.~Cheng, Y.~Zhang, X.~He, W.~Chen, J.~Cheng, and H.~Lu, ``Skeleton-based
  action recognition with shift graph convolutional network,'' in
  \emph{IEEE/CVF Conference on Computer Vision and Pattern Recognition (CVPR)},
  2020, pp. 183--192.

\bibitem{liu2020disentangling}
Z.~Liu, H.~Zhang, Z.~Chen, Z.~Wang, and W.~Ouyang, ``Disentangling and unifying
  graph convolutions for skeleton-based action recognition,'' in \emph{IEEE/CVF
  Conference on Computer Vision and Pattern Recognition (CVPR)}, 2020, pp.
  143--152.

\bibitem{shi2019skeleton}
L.~Shi, Y.~Zhang, J.~Cheng, and H.~Lu, ``Skeleton-based action recognition with
  directed graph neural networks,'' in \emph{IEEE/CVF Conference on Computer
  Vision and Pattern Recognition (CVPR)}, 2019, pp. 7912--7921.

\bibitem{peng2020learning}
W.~Peng, X.~Hong, H.~Chen, and G.~Zhao, ``Learning graph convolutional network
  for skeleton-based human action recognition by neural searching,'' in
  \emph{AAAI Conference on Artificial Intelligence (AAAI)}, 2020, pp.
  2669--2676.

\end{thebibliography}

\vfill

\end{document}